\documentclass[final]{cvpr}

\usepackage{times}
\usepackage{epsfig}
\usepackage{graphicx}
\usepackage{amsmath}
\usepackage{amssymb}
\usepackage{color}

\newif\ifcomments
\commentsfalse %

\usepackage{microtype}
\usepackage{xcolor}
\usepackage[inline]{enumitem}
\usepackage{dsfont}
\usepackage{booktabs}
\usepackage{caption}
\usepackage{subcaption}
\usepackage{multirow}
\usepackage{pifont}
\usepackage{xspace}
\usepackage{amsthm}

\newcommand{\lrc}[1]{\left\{ #1 \right\}}

\newcommand{\sums}[2]{\sum\limits_{#1}^{#2}}
\newcommand{\cl}[1]{\mathcal{#1}}
\newcommand{\T}{^{T}}
\newcommand{\real}{\mathbb{R}}
\newcommand{\indicator}[1]{\mathds{1}_{\{ #1 \}}}

\newcommand{\BB}[1]{\textbf{#1}}

\newcommand{\customparagraph}[1]{\vspace{\parskip}\noindent\textbf{#1}~}
\newcommand{\smalltable}[2][c]{\begin{tabular}{#1}#2\end{tabular}}

\let\exp\undefined
\DeclareMathOperator{\exp}{exp}
\DeclareMathOperator{\relu}{ReLU}
\DeclareMathOperator{\Neg}{Neg}

\newtheorem{proposition}{Proposition}

\def\TRUE{\ding{51}}
\def\FALSE{\ding{53}}
\def\MIS{--}

\def\baseModel{SiMVC\xspace}
\def\baseModelLong{Simple Multi-View Clustering\xspace}

\def\contrastiveModel{CoMVC\xspace}
\def\contrastiveModelLong{Contrastive Multi-View Clustering\xspace}

\usepackage[breaklinks=true,colorlinks,bookmarks=false]{hyperref}

\def\cvprPaperID{8301} %
\def\confYear{CVPR 2021}

\begin{document}

\title{Reconsidering Representation Alignment for Multi-view Clustering}

\author{
  Daniel J. Trosten \hspace{1cm}
  Sigurd Løkse \hspace{1cm}
  Robert Jenssen \hspace{1cm}
  Michael Kampffmeyer\\
  Department of Physics and Technology, UiT The Arctic University of Norway\thanks{UiT Machine Learning Group, \url{machine-learning.uit.no}}\\
}

\maketitle

\begin{abstract}
   
Aligning distributions of view representations is a core component of today's state of the art models for deep multi-view clustering. However, we identify several drawbacks with na{\"i}vely aligning representation distributions. We demonstrate that these drawbacks both lead to less separable clusters in the representation space, and inhibit the model's ability to prioritize views. Based on these observations, we develop a simple baseline model for deep multi-view clustering. Our baseline model avoids representation alignment altogether, while performing similar to, or better than, the current state of the art. We also expand our baseline model by adding a contrastive learning component. This introduces a selective alignment procedure that preserves the model's ability to prioritize views. Our experiments show that the contrastive learning component enhances the baseline model, improving on the current state of the art by a large margin on several datasets\footnote{The source code for the experiments performed in this paper is available at \url{https://github.com/DanielTrosten/mvc}}.

\end{abstract}

\section{Introduction}
  \label{sec:introduction}
  
Several kinds of real world data are gathered from different points of view, or by using a collection of different sensors. Videos, for instance, contain both visual and audible components, while captioned images include both the raw image data and a descriptive text. In both of these examples, the low-level content of the views are vastly different, but they can still carry the same high-level cluster structure. The objective of multi-view clustering is to discover this common clustering structure, by learning from all available views simultaneously.

Learning from multiple sources at once is not a trivial task \cite{baltrusaitisMultimodalMachineLearning2019}. However, the introduction of deep learning \cite{lecunDeepLearning2015a}, has led to the development of several promising deep multi-view clustering models
\cite{abavisaniDeepMultimodalSubspace2018b,liDeepAdversarialMultiview2019,
taoMarginalizedMultiviewEnsemble2020, zhangGeneralizedLatentMultiView2020,
zhouEndtoEndAdversarialAttentionNetwork2020}. These models efficiently learn from multiple views by transforming each view with a view-specific encoder network. The resulting representations are fused to obtain a common representation for all views, which can then be clustered by a subsequent clustering module.

The current state of the art methods for deep multi-view clustering use adversarial training to align the representation distributions from different views \cite{liDeepAdversarialMultiview2019, zhouEndtoEndAdversarialAttentionNetwork2020}.

Aligning distributions leads to view invariant representations, which can be beneficial for the subsequent fusion of views, and the clustering module~\cite{zhouEndtoEndAdversarialAttentionNetwork2020}. View invariant representations preserve the information present in all views, while discarding information that only exists in a subset of views. If the view-specific information is irrelevant to the clustering objective, it will be advantageous for the clustering module that the encoders learn to remove it. Moreover, aligning representation distributions introduces an auxiliary task, which regularizes the encoders, and helps preserve the local geometric structure of the input space. This has been shown to improve single-view deep clustering models \cite{guoImprovedDeepEmbedded2017b}.

Despite these advantages however, we identify three important drawbacks of distribution alignment for multi-view clustering:

\emph{Aligning representations prevents view-prioritization in the representation space.}
Views are not necessarily equally important to the clustering objective. The model should therefore be able to adaptively prioritize views, based on the information contained in the view representations. However, aligning representation distributions makes it harder for the model to prioritize views in the representation space, by making these distributions as similar as possible.

\emph{One-to-one alignment of clusters is only attainable when encoders can separate all clusters in all views.}
When the clustering structure is only partially present in the individual views, alignment causes clusters to merge together in the representation space. This makes the clustering task more difficult for the subsequent clustering module.

\emph{Aligning representation distributions can make it harder to discriminate between clusters.}
Since adversarial alignment only considers the representation distributions, a given cluster from one view might be aligned with a different cluster from another view. This misalignment of label distributions has been shown to have a negative impact on discriminative models in the representation space \cite{zhaoLearningInvariantRepresentation2019}.

The End-to-end Adversarial-attention network for Multi-modal Clustering (EAMC) \cite{zhouEndtoEndAdversarialAttentionNetwork2020} represents the current state of the art for deep multi-view clustering. EAMC aligns the view representations by optimizing an adversarial objective on the encoder networks. The resulting representations are fused with a weighted average, with weights produced by passing the representations through an attention network. Following our reasoning above, we hypothesize that the alignment done by the adversarial module may defeat the purpose of the attention mechanism. Thus inhibiting view prioritization, and leading to less separable clusters after fusion. Our hypothesis is supported by the empirical results of EAMC \cite{zhouEndtoEndAdversarialAttentionNetwork2020}, where the fusion weights are close to uniform for all datasets. Equal fusion weights cause all views to contribute equally to the fused representation, regardless of their contents. Moreover, the fusion weights produced by the attention network depend on all the samples within the current batch. Out-of-sample inference is therefore impossible with EAMC, without making additional modifications to the attention mechanism.

In this work, we seek to alleviate the problems that can arise when aligning distributions of representations in deep multi-view clustering. To this end, we make the following key contributions:
\begin{itemize}%
  \item We highlight pitfalls of aligning representation distributions in deep multi-view clustering, and show that these pitfalls limit previous state of the art models.

  \item We present \baseModelLong (\baseModel) -- a new and simple baseline model for deep multi-view clustering, without any form of alignment. Despite its simplicity compared to existing methods, our experiments show that this baseline model performs similar to -- and in some cases, even better than -- current state of the art methods. \baseModel combines representations of views using a learned linear combination -- a simple but effective mechanism for view-prioritization. We empirically demonstrate that this mechanism allows the model to suppress uninformative views and emphasize views that are important for the clustering objective.

  \item In order to leverage the advantages of alignment -- \ie preservation of local geometric structure, and view invariance -- while simultaneously avoiding the pitfalls, we attach a selective contrastive alignment module to \baseModel. The contrastive module aligns angles between representations at the sample level, circumventing the problem of misaligned label distributions. Furthermore, in the case that one-to-one alignment is not possible, we make the model capable of ignoring the contrastive objective, preserving the model's ability to prioritize views. We refer to this model as \contrastiveModelLong (\contrastiveModel).
\end{itemize}

\section{Pitfalls of distribution alignment in multi-view clustering}
  \label{sec:pitfalls}
  
Here, we consider an idealized version of the multi-view clustering problem. This allows us to investigate and formalize our observations on alignment of representation distributions in multi-view clustering. By assuming that, for each view, all samples within a cluster are located at the same point in the input space, we develop the following proposition\footnote{We provide a proof sketch for Proposition \ref{prop:maxclusters} in the supplementary.}:

\begin{figure*}
  \centering
  \begin{subfigure}[t]{0.245\textwidth}
    \centering
    \includegraphics[height=3.8cm]{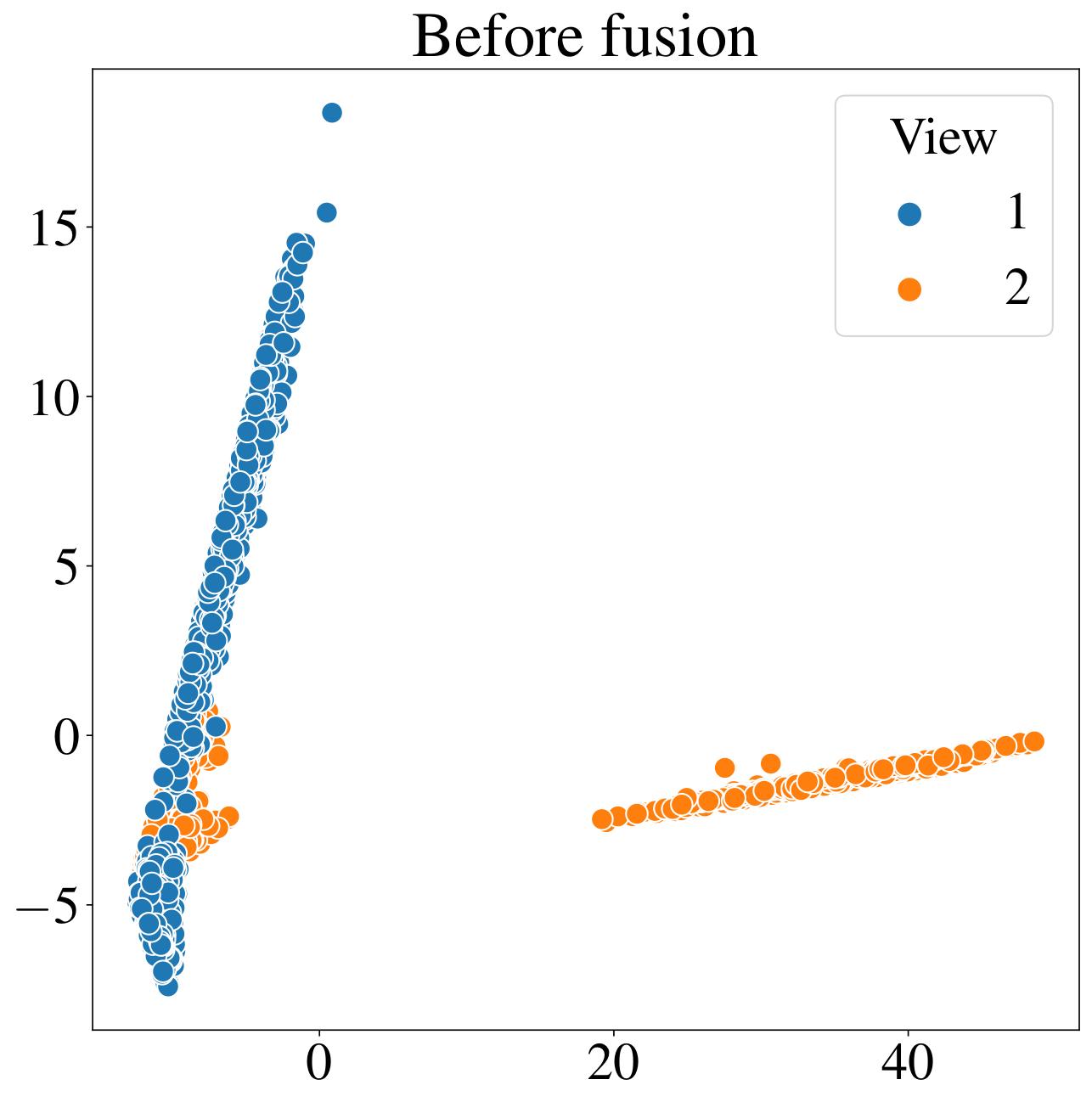}
    \includegraphics[height=3.8cm]{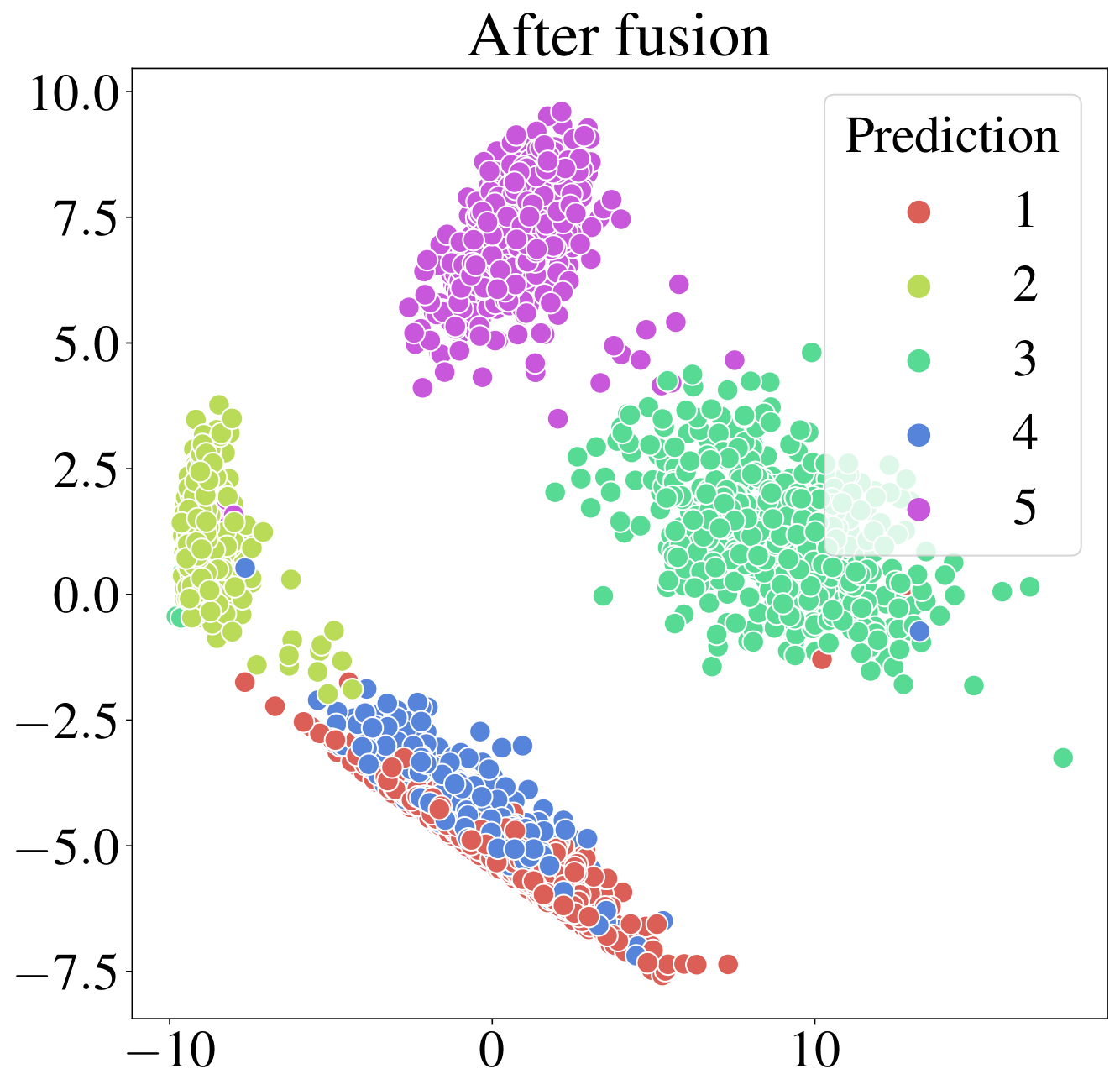}
    \caption{\baseModel+Adv. ACC \( = 0.80 \)}
    \label{fig:toyDataResults_base_adv}
  \end{subfigure}
  \begin{subfigure}[t]{0.245\textwidth}
    \centering
    \includegraphics[height=3.8cm]{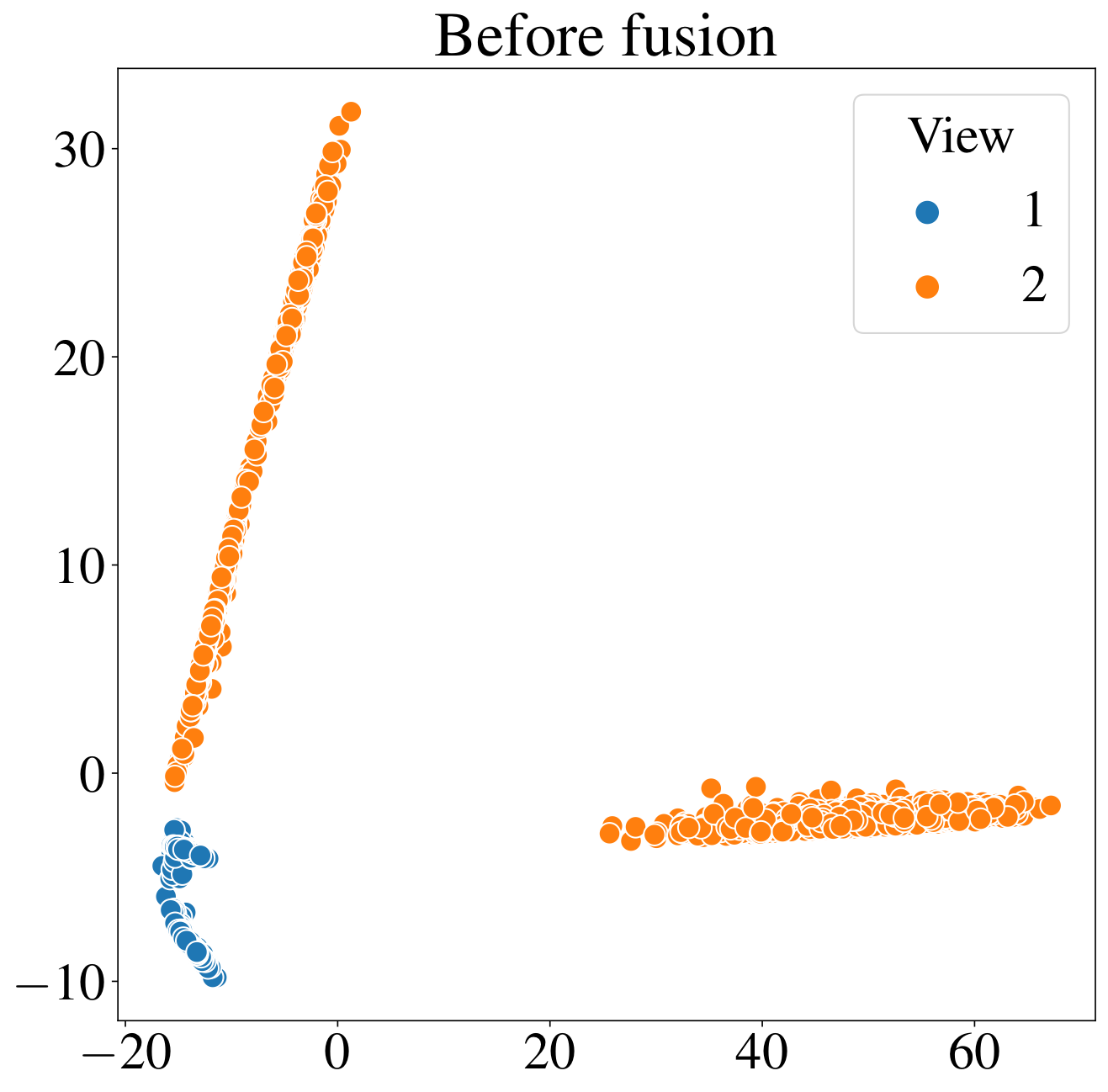}
    \includegraphics[height=3.8cm]{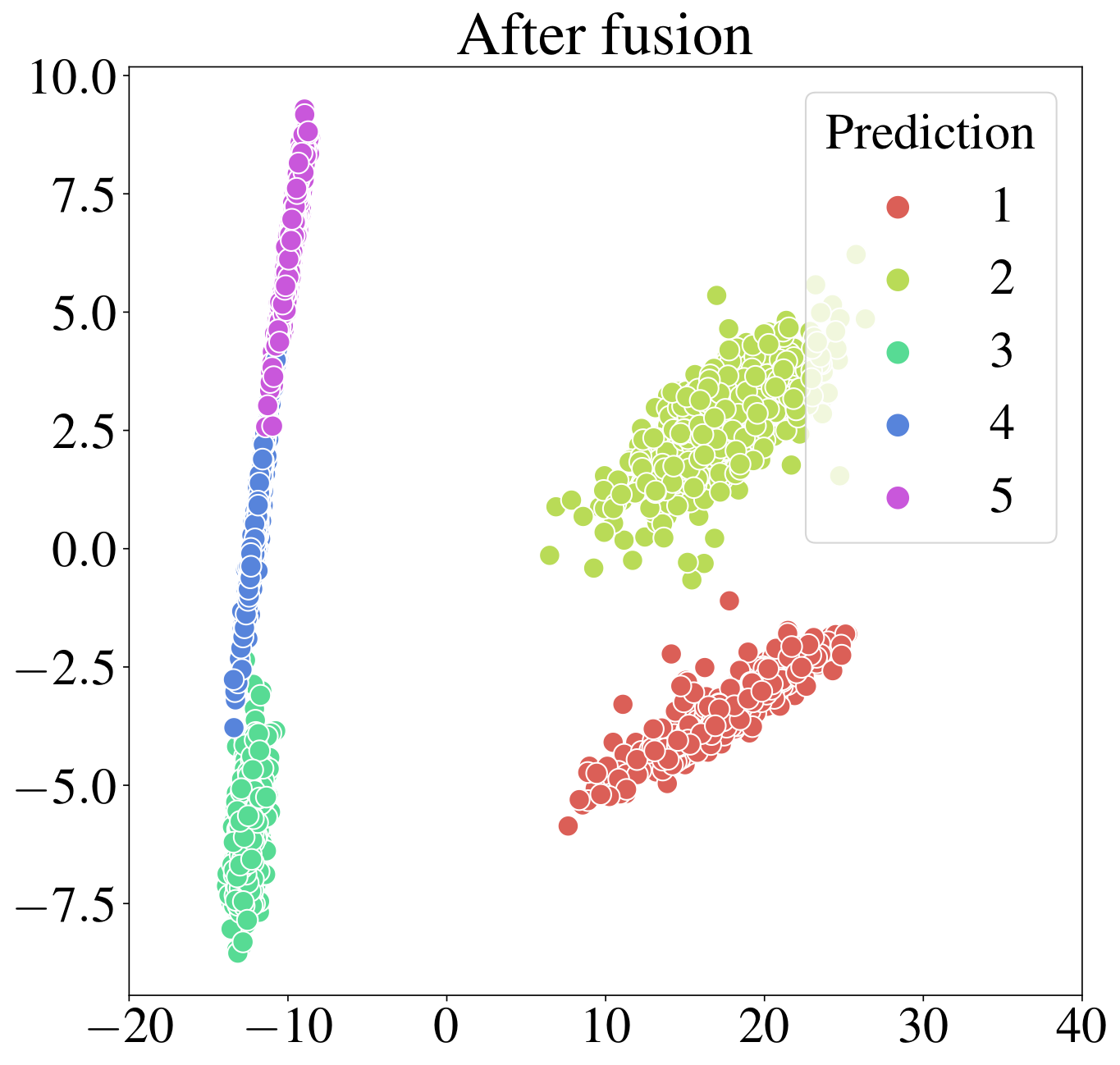}
    \caption{\baseModel. ACC \( = 0.99 \).}
    \label{fig:fig:toyDataResults_base}
  \end{subfigure}
  \begin{subfigure}[t]{0.245\textwidth}
    \centering
    \includegraphics[height=3.8cm]{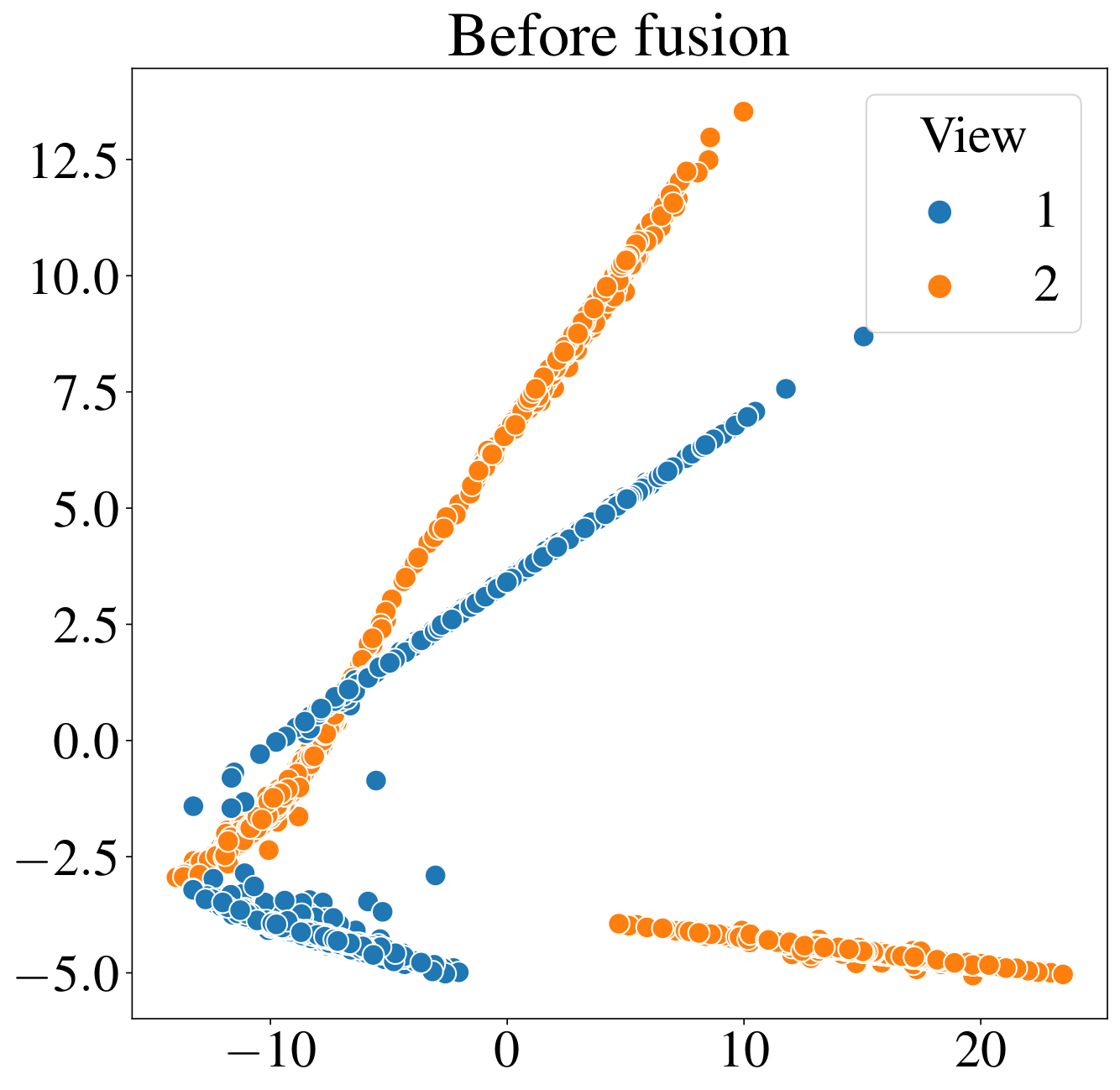}
    \includegraphics[height=3.8cm]{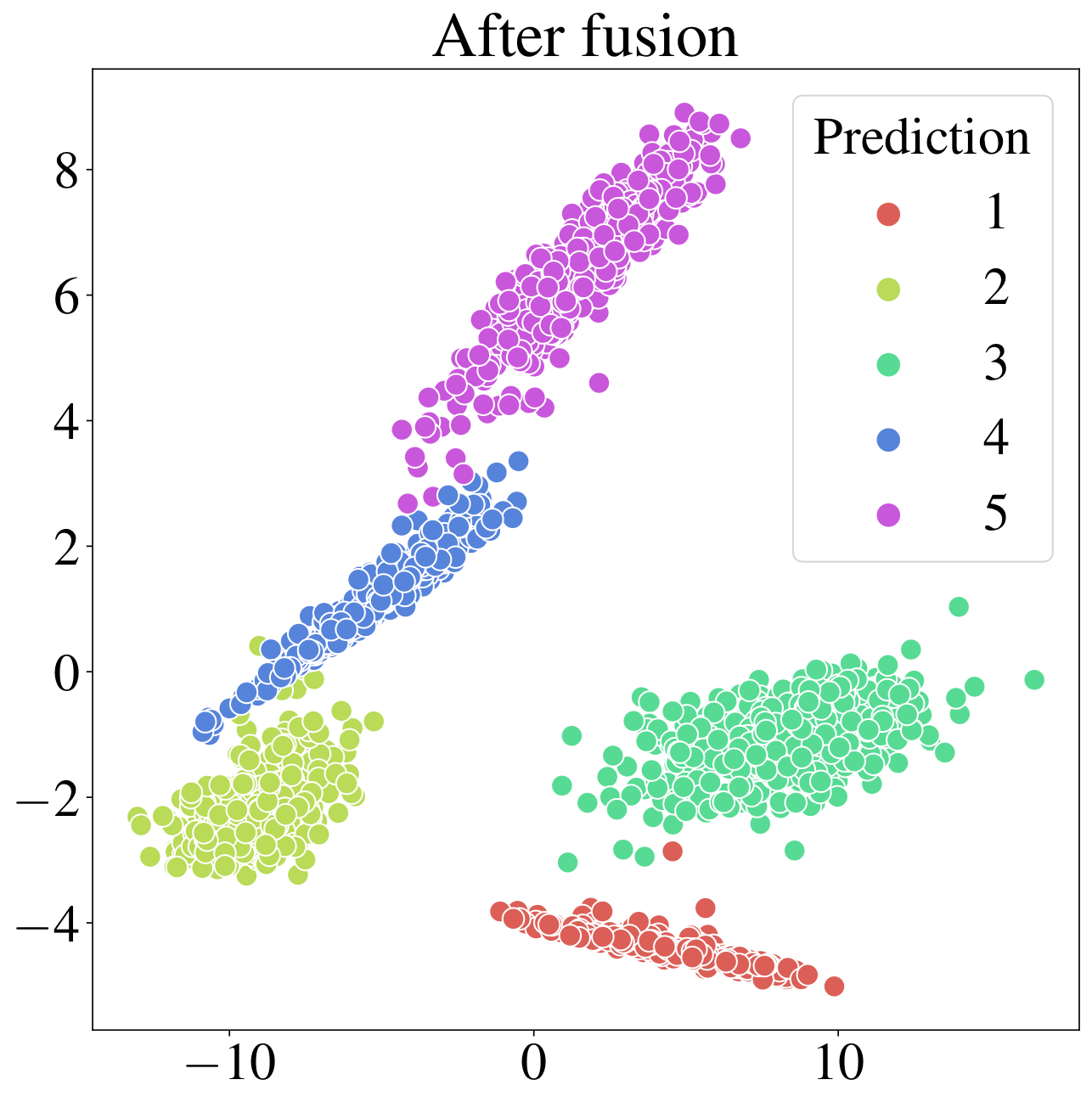}
    \caption{\contrastiveModel. ACC \( = 0.99 \).}
    \label{fig:fig:toyDataResults_contrastive}
  \end{subfigure}
  \begin{subfigure}[t]{0.245\textwidth}
    \centering
    \includegraphics[height=3.8cm]{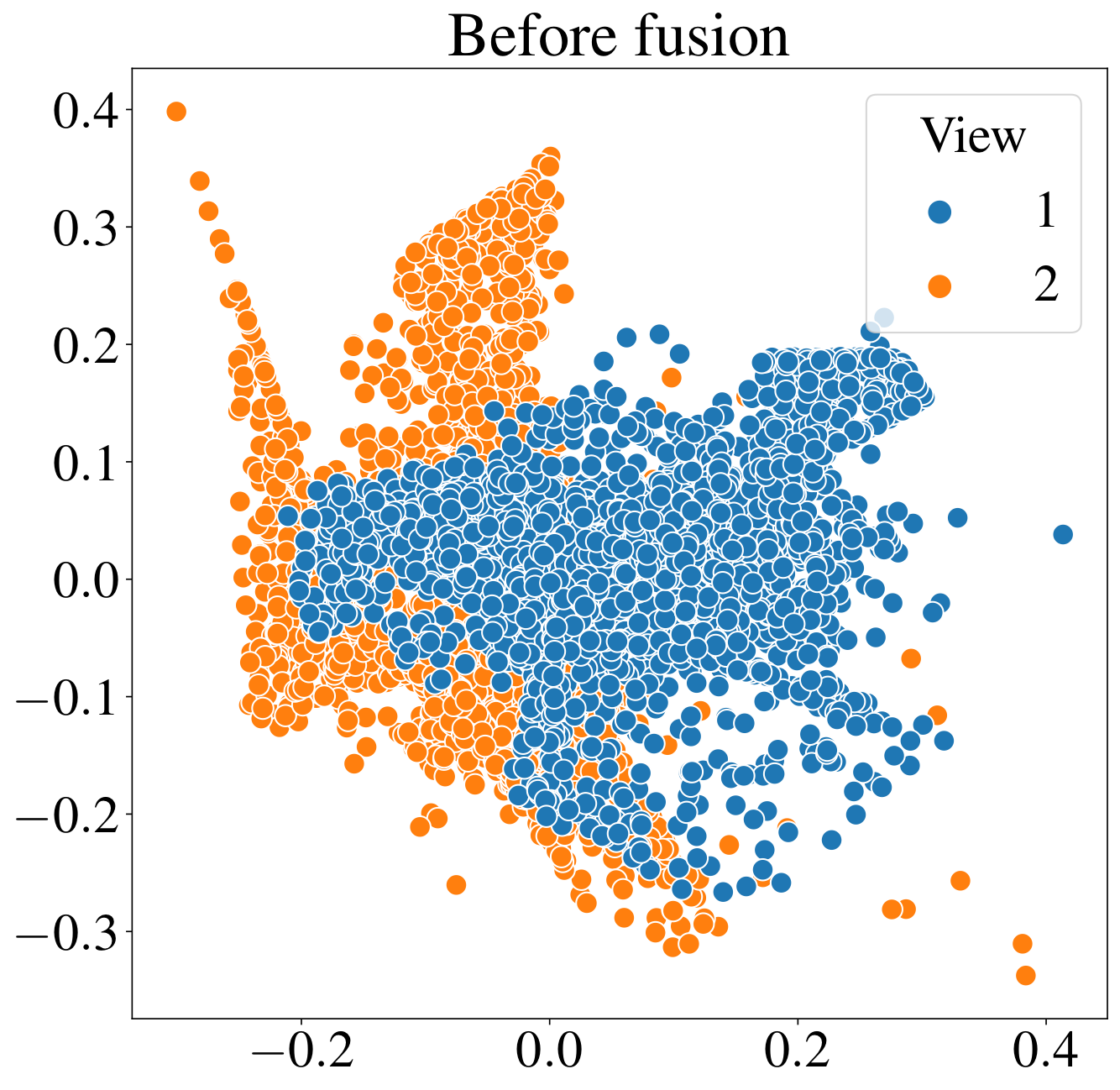}
    \includegraphics[height=3.8cm]{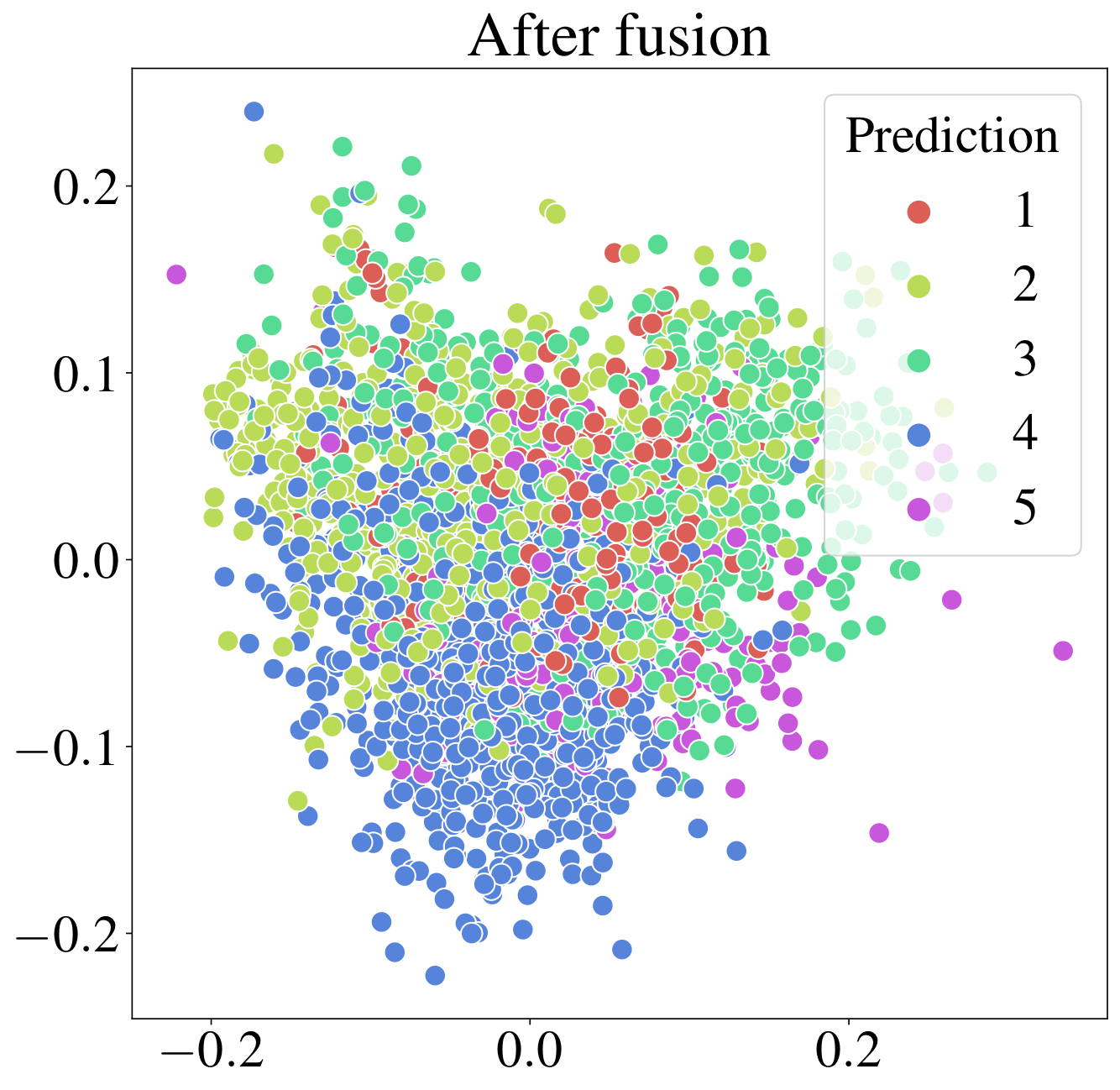}
    \caption{EAMC. ACC \( = 0.368 \).}
    \label{fig:fig:toyDataResults_eamc}
  \end{subfigure}
  \caption{Representations for \baseModel with and without adversarial alignment, \contrastiveModel, and EAMC on our toy dataset.}
  \label{fig:toyDataResults}
\end{figure*}

\begin{figure}
  \centering
  \includegraphics[height=3.9cm]{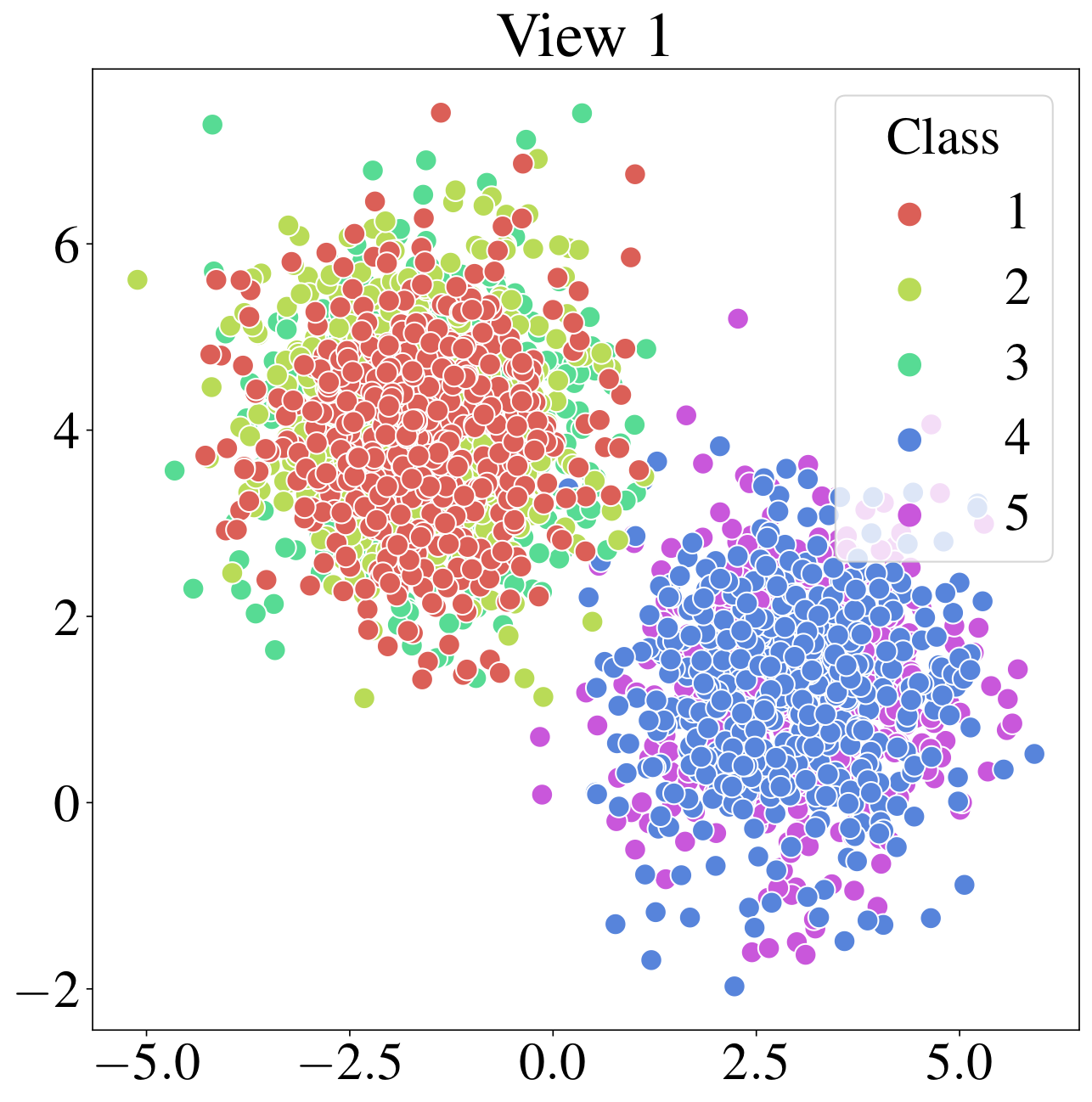}
  \includegraphics[height=3.9cm]{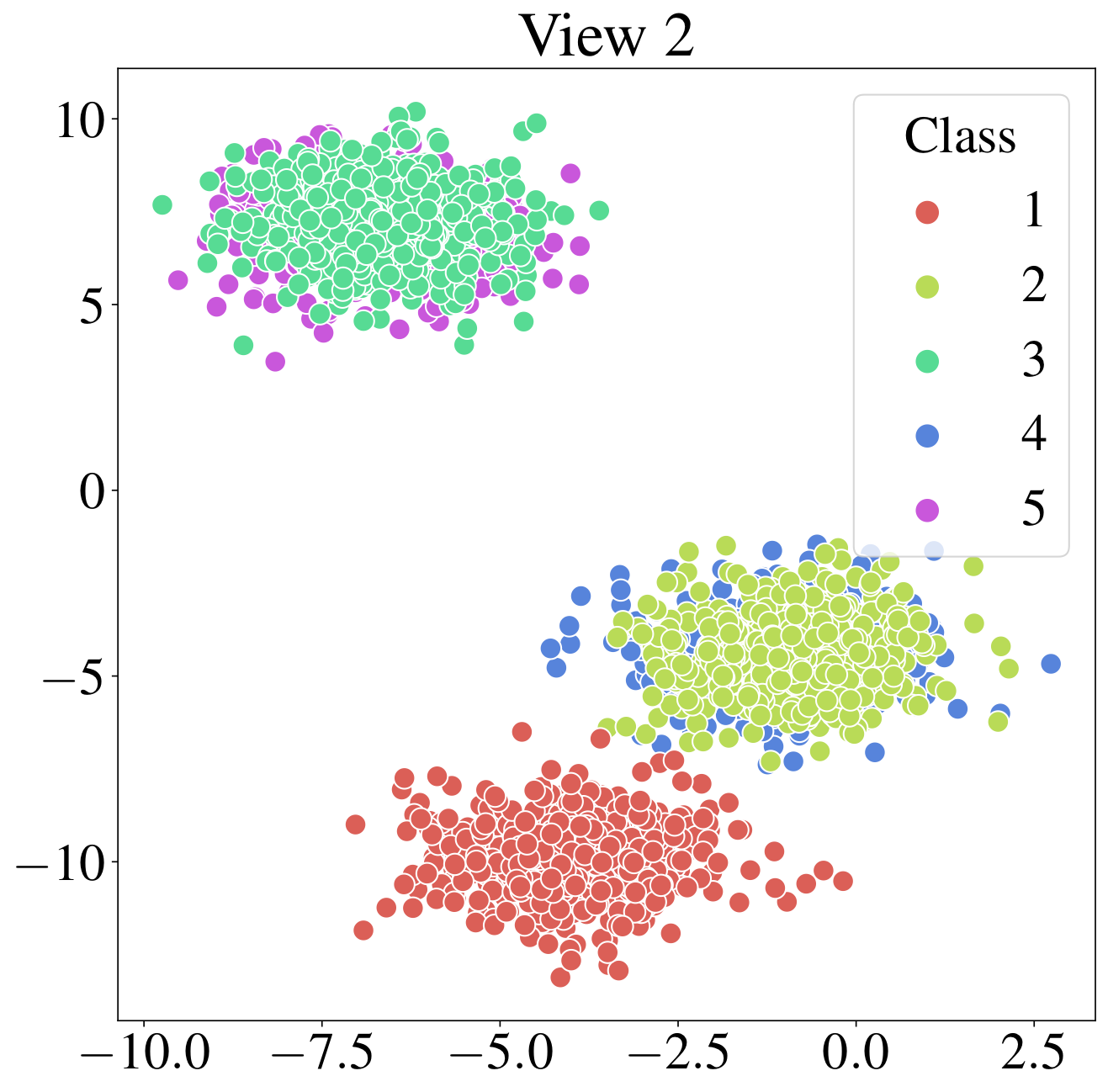}
  \caption{Toy dataset. View 1: Classes (1-3) and (4,5) overlap. View 2: Class 1 is isolated, and classes (2,4) and (3,5) overlap.}
  \label{fig:toydataset}
\end{figure}

\begin{proposition}
  \label{prop:maxclusters}
  Suppose our dataset consists of~\(V\)views and \( k \) ground truth clusters, and we wish to cluster the data according to this ground truth clustering. Furthermore, we make the following assumptions:
  
  \begin{enumerate}[noitemsep]
    \item For each view, all observations that belong to the same ground truth cluster, are located at the same point in the input space.
    \item For a given view \( v \), \( v \in \{1, \dots, V\} \), the number of unique points (\ie distinct/separable clusters) in the input space is \( k_v \).
    \item The views are mapped to representations using view-specific encoders, and subsequently fused according to a linear combination with unique weights.
  \end{enumerate}

  Then, the maximum number of unique clusters after fusion is
  \begin{align}
    \kappa^\text{fused}_\text{aligned} = \min \lrc{ k, \left(\min\limits_{v=1, \dots, V}\{k_v\} \right)^V}
  \end{align}
  if the distributions of representations from different views are perfectly aligned, and
  \begin{align}
    \kappa^\text{fused}_\text{not aligned} = \min\lrc{k, \prod\limits_{v=1}^{V} k_v}
  \end{align}
  if no alignment is performed.
\end{proposition}

\customparagraph{Implications of Proposition \ref{prop:maxclusters}.}
\( \kappa^\text{fused}_\cdot \) in Proposition \ref{prop:maxclusters} controls how well the clustering module is able to cluster the fused representations. If \( \kappa^\text{fused}_\cdot < k \), it means that some of the clusters are located at the same point after fusion, making it impossible for the clustering module to discriminate between these clusters. In the extreme case that one of the views groups all the clusters together (\ie \( k_v = 1 \)), it follows that \( \kappa^\text{fused}_\text{aligned} = 1 \). This happens because all other views are aligned to the uninformative view (for which \( k_v = 1 \)), collapsing the cluster structure in the representation space. Alignment thus prevents the suppression of this view, and makes it harder to discriminate between clusters in the representation space.

However, if we are able to discriminate between all clusters in all views, we have \( k_v = k \) for all views, resulting in
\( \kappa^\text{fused}_\text{aligned} = \kappa^\text{fused}_\text{not aligned} = k \).
In this case it is possible for both alignment-based models and non-alignment-based models to perfectly cluster the data, provided that the clustering module is sufficiently capable. Alignment-based models can thus benefit from the advantages of alignment, while still being able to separate clusters after fusion.

\customparagraph{Experiments on toy data.}
Proposition \ref{prop:maxclusters} makes the simplification that all samples within a cluster are located at the same point, for each view. In order to demonstrate the potential negative impact of aligning representation distributions in a less idealistic setting, and to further motivate the problem, we create a simple two-view dataset. The dataset is shown in Figure \ref{fig:toydataset}, and contains five elliptical clusters in two two-dimensional views\footnote{We repeat this experiment for a \( 3 \)-cluster dataset in the supplementary.}.

We fit \baseModel and \baseModel with adversarial alignment (\baseModel+Adv.) to this dataset, in order to demonstrate the effects of aligning distributions, in a controlled setting. Additionally, we fit our \contrastiveModel and the current state of the art, EAMC, to evaluate more advanced alignment procedures. Note that, for all of these models, the fusion is implemented as a weighted average of view representations, as in Proposition \ref{prop:maxclusters}. The remaining details on \baseModel and \contrastiveModel are provided in the next section.

Figures \ref{fig:toyDataResults_base_adv} and \ref{fig:fig:toyDataResults_base} show that attempting to align distributions with adversarial alignment prevents \baseModel from separating between clusters \( 1 \) and \( 4 \). By adding the adversarial alignment to \baseModel, the number of visible clusters after fusion is reduced from \( 5 \) to \( 4 \). This is in line with Proposition \ref{prop:maxclusters}, since we have \( \kappa^\text{fused}_\text{aligned} = 4 \) and \( \kappa^\text{fused}_\text{not aligned} = 5 \) for this dataset.
Figure \ref{fig:fig:toyDataResults_contrastive} shows that \contrastiveModel, which relies on the cosine similarity, aligns the angles between the majority of observations. This alignment does not cause classes to overlap in the fused representation. EAMC attempts to align the distributions of view representations (Figure \ref{fig:fig:toyDataResults_eamc}), resulting in a fused representation where the classes are hard to separate. Interestingly, the resulting fused representation exhibits a single group of points, which is significantly worse than the upper bound \( \kappa^\text{fused}_\text{aligned} = 4 \) in the analogous idealistic setting. We hypothesize that this is due to EAMC's fusion weights, which we observed to be almost equal for this experiment -- thus breaking assumption 3 in Proposition \ref{prop:maxclusters}.

\section{Methods}
  \label{sec:method}
  
\subsection{\baseModelLong (\baseModel)}
  \label{sec:method_base}
  Suppose our dataset consists of \( n \) objects observed from \( V \) views. Let \( x_i^{(v)} \) be the observation of object \( i \) from view \( v \). The objective of our models is then to assign the set of views for each object, \( \{ x_i^{(1)}, \dots, x_i^{(V)} \} \), to one of \( k \) clusters.

  To achieve this, we first transform each \( x_i^{(v)} \) to its representation \( z_i^{(v)} \) according to
  \begin{align}
    \label{eq:representation}
    z_i^{(v)} = f^{(v)}(x_i^{(v)})
  \end{align}
  where \( f^{(v)} \) denotes the encoder network for view \( v \). We then compute the fused representation as a weighted average
  \begin{align}
    z_i = \sums{v=1}{V} w_v z_i^{(v)}
  \end{align}
  where \( w_1, \dots, w_v \) are the \emph{fusion weights}, satisfying \( w_v > 0 \) for \( v = 1, \dots, V \) and \( \sum_{v=1}^{V} w_v = 1 \). We enforce these constraints by keeping a set of unnormalized weights, from which we obtain \( w_1, \dots, w_V \) using the softmax function. We let the unnormalized weights be trainable parameters -- a design choice which has the following advantages:
  \begin{enumerate*}[label=(\roman*)]
    \item During training, the model has a simple and interpretable way to prioritize views according to its clustering objective. By not relying on an auxiliary attention network, we also make the model more efficient -- both in terms of memory consumption and training time\footnote{Average training times for \baseModel, \contrastiveModel, and EAMC are given in the supplementary.}.
    \item In inference, the weights act as any other model parameters, meaning that out-of sample inference can be done with arbitrary batch sizes, without any modifications to the trained model. Fixed fusion weights also result in deterministic predictions, which are independent of any other samples within the batch.
  \end{enumerate*}

  To obtain the final cluster assignments, we pass the fused representation through a fully connected layer, producing the hidden representation \( h_i \). This is processed by another fully connected layer with a softmax activation, to obtain the \( k \)-dimensional vector of soft cluster assignments, \( \alpha_i \).

  \customparagraph{Loss function.}
  We adopt the Deep Divergence-based Clustering (DDC) \cite{kampffmeyerDeepDivergencebasedApproach2019} loss, which has shown state of the art performance in single-view image clustering \cite{kampffmeyerDeepDivergencebasedApproach2019}. This is also the clustering loss used by EAMC \cite{zhouEndtoEndAdversarialAttentionNetwork2020} -- the current state of the art method for multi-view clustering.

  The clustering loss consists of three terms, enforcing cluster separability and compactness, orthogonal cluster assignments, and closeness of cluster assignments to simplex corners, respectively.
  The first loss term is derived from the multiple-density generalization of the Cauchy-Schwarz divergence \cite{jenssenCauchySchwarzDivergence2006a}, and requires clusters to be separable and compact in the space of hidden representations:
  \begin{align}
    \cl L_1 = \sums{i=1}{k-1} \sums{j=i+1}{k} \frac{ \binom{k}{2}^{-1} \sums{a=1}{n} \sums{b=1}{n} \alpha_{ai} \kappa_{ab} \alpha_{bj}}{ \sqrt{ \sums{a=1}{n}\sums{b=1}{n} \alpha_{ai} \kappa_{ab} \alpha_{bi} \sums{a=1}{n}\sums{b=1}{n} \alpha_{aj} \kappa_{ab} \alpha_{bj} } }
  \end{align}
  where \( k \) denotes the number of clusters, \( \kappa_{ij} = \exp(-||h_i - h_j||^2/(2 \sigma^2)) \), and \( \sigma \) is a hyperparameter.

  The second term encourages the cluster assignment vectors for different objects to be orthogonal:
  \begin{align}
    \cl L_2 = \binom{n}{2}^{-1} \sums{i=1}{n-1} \sums{j=i+1}{n} \alpha_i\T \alpha_j.
  \end{align}
  Finally, the third term pushes the cluster assignment vectors close to the standard simplex in \( \real^k \):
  \begin{align}
    \cl L_3 = \sums{i=1}{k-1} \sums{j=i+1}{k} \frac{ \binom{k}{2}^{-1} \sums{a=1}{n} \sums{b=1}{n} m_{ai} \kappa_{ab} m_{bj}}{ \sqrt{ \sums{a=1}{n}\sums{b=1}{n} m_{ai} \kappa_{ab} m_{bi} \sums{a=1}{n}\sums{b=1}{n} m_{aj} \kappa_{ab} m_{bj} } }
  \end{align}
  where \( m_{ij} = \exp(-||\alpha_i - e_j||^2) \), and \( e_j \) is corner \( j \) of the standard simplex in \( \real^k \).

  The final clustering loss which we minimize during training of \baseModel is the sum of these three terms:
  \begin{align}
    \label{eq:loss_cluster}
    \cl L_\text{cluster} = \cl L_1 + \cl L_2 + \cl L_3.
  \end{align}

\subsection{\contrastiveModelLong (\contrastiveModel)}
  \label{sec:method_contrastive}
  \begin{figure}
    \centering
    \includegraphics[width=0.9\columnwidth]{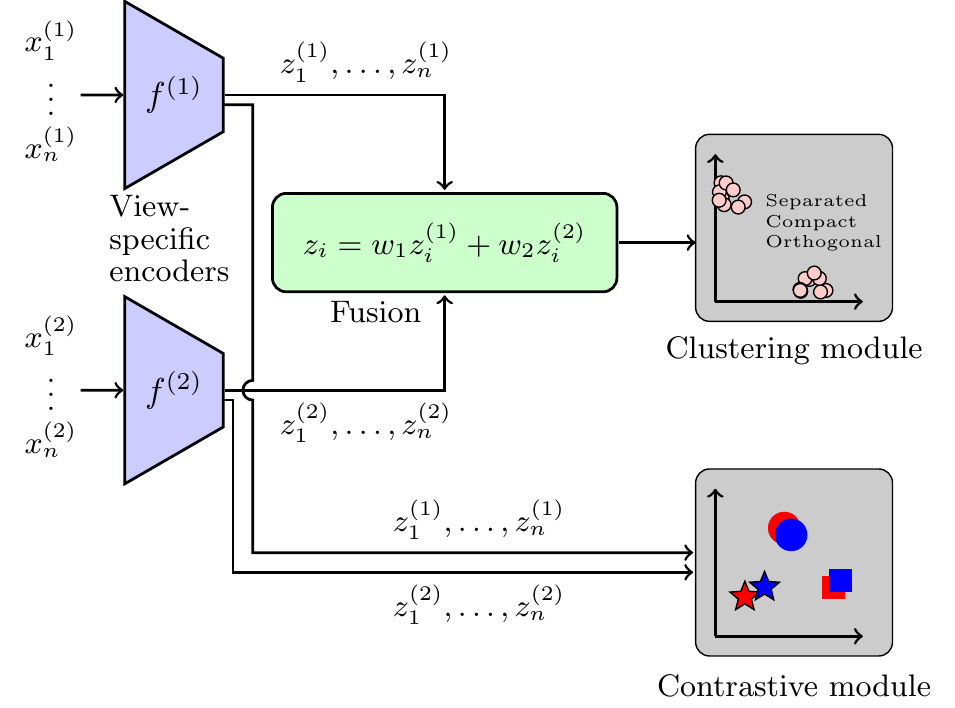}
    \caption{Overview of our proposed models for a two-view dataset. In both \baseModel and \contrastiveModel, the views are first encoded by the view-specific encoder networks \( f^{(1)} \) and \( f^{(2)} \). The resulting representations are fused with a weighted mean, and then clustered by the clustering module. \contrastiveModel includes an additional contrastive module.}%
    \label{fig:model}
  \end{figure}
  Contrastive learning offers a way to align representations from different views at the sample level, forcing the label distributions to be aligned as well. Our hypothesis is therefore that a \emph{selective} contrastive alignment will allow the model to learn common representations that are well suited for clustering -- while simultaneously avoiding the previously discussed pitfalls of distribution alignment.
  Self-supervised contrastive models have shown great potential for a large variety of downstream computer vision tasks
  \cite{bachmanLearningRepresentationsMaximizing2019,chenSimpleFrameworkContrastive2020a,
  chenBigSelfSupervisedModels2020,grillBootstrapYourOwn2020,heMomentumContrastUnsupervised2020,
  lowePuttingEndEndtoEnd2019,oordRepresentationLearningContrastive2019}.
  These models learn image representations by requiring that representations from \emph{positive} pairs are mapped close together, while representations from \emph{negative} pairs are mapped sufficiently far apart.
  In multi-view learning, each object has a set of observations from different views associated with it. This admits a natural definition of pairs: Let views of the \emph{same} object be positive pairs, and views of \emph{different} objects be negative pairs.

  Following \cite{chenSimpleFrameworkContrastive2020a}, we compute the similarity of two representations \( z_i^{(v)} \) and \( z_j^{(u)} \) as the cosine similarity:
  \begin{align}
    s^{(vu)}_{ij} = \frac{z_i^{(v)}{\T} z_j^{(u)} }{|| z_i^{(v)} || \cdot || z_j^{(u)} ||}.
  \end{align}
  Note that in \cite{chenSimpleFrameworkContrastive2020a}, they show that the addition of a projection head between the representations and the similarity, results in better representations -- in terms of linear classification accuracy on the learned representations. We found that this was not the case for our model, so we chose to compute the similarity on the representations directly. Experiments comparing versions of our model with and without the projection head can be found in the supplementary.

  In order to define a contrastive loss for an arbitrary number of views, we introduce the following generalized version of the NT-Xent loss \cite{chenSimpleFrameworkContrastive2020a}:
  \begin{align}
     \cl L_\text{contrastive} = \frac{1}{n V(V-1)} \sums{i=1}{n} \sums{v=1}{V} \sums{u=1}{V} \indicator{u \neq v} \ell_i^{(uv)}
  \end{align}
  where \( \indicator{u \neq v} \) evaluates to \( 1 \) when \( u \neq v \) and \( 0 \) otherwise, and
  \begin{align}
    \label{eq:contrasive}
    \ell_{i}^{(uv)} = - \log \frac{ e^{s_{ii}^{(uv)} / \tau} }{ \sum_{s' \in \Neg(z_i^{(v)}, z_i^{(u)})} e^{s' / \tau}}.
  \end{align}
  Here, \( \tau \) is a hyperparameter\footnote{We set \( \tau = 0.1 \) for all experiments, following \cite{chenSimpleFrameworkContrastive2020a}.}, and \( \Neg(z_i^{(v)}, z_i^{(u)}) \) denotes the set of similarities for negative pairs corresponding to the positive pair \( (z_i^{(v)}, z_i^{(u)}) \).

  A straightforward way to construct \( \Neg(z_i^{(v)}, z_i^{(u)}) \) would be to include the similarity between all views of object \( i \), and all views of all the other objects within the current batch. However, minimizing Eq. \eqref{eq:contrasive} will result in negative samples having a low similarity score. This is indeed the objective of ordinary contrastive learning, but it might be counteractive to the clustering objective, where we want objects from the same cluster to be grouped together in the representation space, and thus be similar to each other. To prevent the contrastive loss from breaking this group structure, we construct \( \Neg(z_i^{(v)}, z_i^{(u)}) \) in the following manner: First, we define the set
  \begin{align}
    \nonumber
    \cl N_i = \{ s_{ij}^{(uv)} : \ & j = 1, \dots, n,\ j \neq i, u, v = 1, \dots, V, \\
    & \arg\max \alpha_i \neq \arg\max \alpha_j \},
  \end{align}
  which consists of all similarities between all views of object \( i \), and all views of all other objects \emph{that were assigned to a different cluster than object} \( i \). We then construct \( \Neg(z_i^{(v)}, z_i^{(u)}) \) by sampling a fixed number of similarities from \( \cl N_i \). This procedure ensures that we only repel representations of objects that were assigned to different clusters by the clustering module.

  \contrastiveModel is the result of adding this contrastive learning framework to \baseModel. Figure \ref{fig:model} shows a schematic overview of the model for a dataset containing two views.

  The loss we use to train \contrastiveModel is
  \begin{align}
    \cl L = \cl L_\text{cluster} + \delta \cdot \min \{w_1, \dots, w_V \} \cl L_\text{contrastive}
  \end{align}
  where \( \cl L_\text{cluster} \) is the clustering loss defined in Eq. \eqref{eq:loss_cluster}, and \( \delta \) is a hyperparameter which influences the strength of the contrastive loss. \( w_1, \dots w_V \) are the fusion weights from \baseModel\footnote{Note that we do not propagate gradients through the \( \min \) operation, in order to avoid the trivial solution of setting the smallest fusion weight to \( 0 \).}.

  Minimizing the contrastive loss results in representations that have high cosine similarities. The contrastive alignment is therefore
  \begin{enumerate*}[label=(\roman*)]
    \item \emph{approximate}, since only the angles between representations, and not the representations themselves, are considered; and
    \item \emph{at the sample level}, preventing misaligned label distributions.
  \end{enumerate*}
  Furthermore, multiplying the contrastive loss with the smallest fusion weight automatically adjusts the strength of the contrastive loss, according to the weight of the least informative view. The alignment is therefore \emph{selective}: If the model learns to discard a view by setting its fusion weight to \( 0 \), it will simultaneously disable the alignment procedure.
  By adapting the alignment weight and not relying on adversarial training, \contrastiveModel can benefit from the advantages of aligning representations, while circumventing both the drawbacks of adversarial alignment, and possible difficulties with min-max optimization \cite{arjovskyWassersteinGenerativeAdversarial2017, goodfellowGenerativeAdversarialNets2014a}.

\section{Related work}
  \label{sec:relatedWork}

In this section we will give a brief summary of the existing work on multi-view clustering, as well as related work discussing modality alignment in multi-modal learning. Existing methods for multi-view clustering can be divided into two categories: Traditional (non-deep learning based) methods and deep learning based methods.

\customparagraph{Traditional methods.}
Two-stage methods first learn a common representation from all the views, before clustering them using a single-view clustering algorithm~\cite{blaschkoCorrelationalSpectralClustering2008,chaudhuriMultiViewClusteringCanonical2009}. However, recent work shows that letting the learned representation adapt to the clustering algorithm leads to better clusterings \cite{xieUnsupervisedDeepEmbedding2016a}. In order to avoid this drawback of two-stage approaches, non-negative matrix factorization
\cite{caiMultiViewKMeansClustering2013, chengTensorBasedLowDimensionalRepresentation2019, huangMultiViewDataFusion2020, xuMultiviewSelfPacedLearning2015, zhaoMultiViewClusteringDeep2017} has been used to compute the cluster assignment matrix directly from the data matrices. Similarly, subspace methods assume that observations can be represented by one or more self-representation matrices
\cite{bachmanLearningRepresentationsMaximizing2019,
caoDiversityinducedMultiviewSubspace2015, liuRobustRecoverySubspace2013,
luoConsistentSpecificMultiView2018,xieMultiviewClusteringJoint2019,yangSplitMultiplicativeMultiView2019,
 yinMultiviewSubspaceClustering2019,zhangGeneralizedLatentMultiView2020} and use the self-representation matrices to identify linear subspaces of the vector space spanned by all the observations, that represent distinct clusters. Alternative popular approaches include methods based on graphs
\cite{nieSelfweightedMultiviewClustering2017, taoMarginalizedMultiviewEnsemble2020, wangRobustSelfWeightedMultiView2020, wangParameterFreeWeightedMultiView2020, zhanMultiviewConsensusGraph2019, zongWeightedMultiViewSpectral2018}
and kernels
\cite{duRobustMultipleKernel2015,gonenLocalizedDataFusion2014,liMultipleKernelClustering2016,
liuMultipleKernelKMeans2016}, which both assume that the data can be represented with one or more kernel (or affinity) matrices such that respective clusterings can be found based on these matrices.

\customparagraph{Deep learning based methods.}
Deep learning based two-stage methods \cite{andrewDeepCanonicalCorrelation2013, ngiamMultimodalDeepLearning2011,
wangDeepMultiViewRepresentation2015} work similarly to the two-stage methods described above, but instead use deep neural networks to learn the common representation.
However, the two-stage methods are regularly outperformed by deep end-to-end methods that adapt their representation learning networks to the subsequent clustering module. Deep graph-based methods \cite{chengMultiViewAttributeGraph2020,huangAutoweightedMultiviewClustering2020,huangMultiSpectralNetSpectralClustering2019,
liDeepGraphRegularized2020} for instance, use affinity matrices together with graph neural networks to directly cluster the data. Similarly, deep subspace methods \cite{abavisaniDeepMultimodalSubspace2018b,araujoSelforganizingSubspaceClustering2020}
make the same subspace assumption as above, but compute the self-representation matrix from an intermediate representation in their deep neural network. Lastly, adversarial methods \cite{liDeepAdversarialMultiview2019,zhouEndtoEndAdversarialAttentionNetwork2020} use generators and discriminators to align distributions of hidden representations from different views.
These adversarial methods have outperformed the previous approaches to multi-view clustering, yielding state of the art clustering performance on several multi-view datasets.

\customparagraph{Distribution alignment.} Outside the field of multi-view clustering, the problem of na{\"i}vely aligning distributions has recently found increasing attention \cite{zhaoLearningInvariantRepresentation2019, wu2019domain}, and led to more efficient fusion techniques \cite{hou2019deep, cangea2019xflow, perez2019mfas}. However, this effort has largely been restricted to supervised multi-modal learning frameworks and domain adaptation approaches.

\section{Experiments}
  \label{sec:experiments}
  
\subsection{Setup}
  \customparagraph{Implementation.}
    Our models are implemented in the PyTorch \cite{paszkePytorch2019} framework. We train the models for \( 100 \) epochs on mini-batches of size \( 100 \), using the Adam optimization technique \cite{kingmaAdamMethodStochastic2015} with default parameters. We observe that \( 100 \) epochs is sufficient for the training to converge. Training is repeated \( 20 \) times, and we report the results from the run resulting in the lowest value of \( \cl L_1 \) in the clustering loss. The \( \sigma \) hyperparameter was set to \( 15 \% \) of the median pairwise distance between hidden representations within a mini-batch, following
    \cite{kampffmeyerDeepDivergencebasedApproach2019}. For the contrastive model, we set the number of negative pairs per positive pair to \( 25 \) for all experiments. We set \( \delta = 0.1 \) for the two-view datasets, and \( \delta = 20 \) for the three-view datasets. We observe that the three-view datasets benefit from stronger contrastive alignment. Our implementation and a complete overview of the architecture details can be found in the supplementary.

  \customparagraph{Datasets.}
  We evaluate our models using six well-known multi-view datasets \cite{liDeepAdversarialMultiview2019,zhouEndtoEndAdversarialAttentionNetwork2020}, containing both raw image data, and vector data. These are:
  \begin{enumerate*}[label=(\roman*)]
    \item \emph{PASCAL VOC 2007 (VOC)} \cite{everinghamVOC2010}. We use the version provided by \cite{hwangAccounting2010}, which contains GIST features and word frequency counts for manually tagged natural images.
    \item \emph{Columbia Consumer Video (CCV)} \cite{jiangConsumerVideoUnderstanding2011}, which consists of SIFT, STIP and MFCC features from internet videos.
    \item \emph{Edge-MNIST (E-MNIST)} \cite{liuCoupledGenerativeAdversarial2016}, which is a version of the ordinary MNIST dataset where the views contain the original digit, and an edge-detected version, respectively.
    \item \emph{Edge-FashionMNIST (E-FMNIST)} \cite{xiaoFashionMNISTNovelImage2017a}, which consists of grayscale images of clothing items. We synthesize a second view by running the same edge-detector as the one used to create E-MNIST.
    \item \emph{COIL-20} \cite{neneColumbiaObjectImage1996b}, which contains grayscale images of \( 20 \) items, depicted from different angles. We create a three-view dataset by randomly grouping the images for an item into groups of three.
    \item \emph{SentencesNYU v2 (RGB-D)} \cite{kongWhatAreYou2014}, which consists of images of indoor scenes along with descriptions of each image. Following \cite{zhouEndtoEndAdversarialAttentionNetwork2020}, we use image features from a ResNet-50 without the classification head, pre-trained on the ImageNet dataset, as the first view. Embeddings of the image descriptions using a pre-trained doc2vec model on the Wikipedia dataset constitute the second view\footnote{We provide the details of these pre-trained models in the supplementary.}.
  \end{enumerate*}

  Note that, for the datasets with multiple labels, we select the objects with exactly one label. See Table \ref{tab:datasets} for more information on the evaluation datasets.

  \begin{table}
    \centering
    
\bgroup
\def\arraystretch{0.85}
\setlength{\tabcolsep}{4pt}
\begin{tabular}{lcccc} \toprule
  Dataset  & Objs.       & Cats.    & Views & Dims. \\ \midrule
  VOC      & \( 5649 \)  & \( 20 \) & \( 2 \) & \( 512,\ 399 \) \\
  CCV      & \( 6773 \)  & \( 20 \) & \( 3 \) & \( 5000,\ 5000,\ 4000 \) \\
  E-MNIST  & \( 60000 \) & \( 10 \) & \( 2 \) & \( 28 \times 28 \) \\
  E-FMNIST & \( 60000 \) & \( 10 \) & \( 2 \) & \( 28 \times 28 \) \\
  COIL-20  & \( 480 \)   & \( 20 \) & \( 3 \) & \( 128 \times 128 \) \\
  RGB-D    & \( 1449 \)  & \( 13 \) & \( 2 \) & \( 2048,\ 300 \) \\
  \bottomrule
\end{tabular}

\egroup

    \caption{Summary of the datasets used for evaluation. \emph{Objs.}\ and \emph{Cats.}\ denote the number of objects and categories present in the dataset, respectively. \emph{Views} and \emph{Dims.}\ denote the number of views, and the dimensionality of each view, respectively. Note that for E-MNIST, E-FMNIST, and COIL-20, the input dimensionality is the same for all views.}
    \label{tab:datasets}
  \end{table}

  \customparagraph{Baseline models.}
  We compare our models to an extensive set of baseline methods, which represent the current state of the art for multi-view clustering:
  \begin{enumerate*}[label=(\roman*)]
    \item Spectral Clustering (SC) \cite{shiNormalizedCutsImage2000} on each view, and the concatenation of all views SC(con);
    \item Robust Multi-view K-means Clustering (RMKMC) \cite{caiMultiViewKMeansClustering2013};
    \item tensor-based Representation Learning Multi-view clustering tRLMvc \cite{chengTensorBasedLowDimensionalRepresentation2019};
    \item Consistent and Specific Multi-view Subspace Clustering (CSMSC) \cite{luoConsistentSpecificMultiView2018};
    \item Weighted Multi-view Spectral Clustering (WMSC) \cite{zongWeightedMultiViewSpectral2018};
    \item Multi-view Consensus Graph Clustering (MCGC) \cite{zhanMultiviewConsensusGraph2019};
    \item Deep Canonical Correlation Analysis (DCCA) \cite{andrewDeepCanonicalCorrelation2013};
    \item Deep Multimodal Subspace Clustering (DMSC) \cite{abavisaniDeepMultimodalSubspace2018b};
    \item Deep Adversarial Multi-view Clustering (DAMC) \cite{liDeepAdversarialMultiview2019}; and
    \item End-to-end Adversarial attention network for Multi-modal Clustering (EAMC) \cite{zhouEndtoEndAdversarialAttentionNetwork2020}.
  \end{enumerate*}

  \customparagraph{Evaluation protocol.}
  To ensure a fair comparison, we report the baseline results over multiple runs, following \cite{zhouEndtoEndAdversarialAttentionNetwork2020}\footnote{The details of the evaluation protocol are given in the supplementary.}. To assess the models' clustering performance, we use the unsupervised clustering accuracy (ACC) and normalized mutual information (NMI). For both these metrics, higher values correspond to better clusterings.

\subsection{Results}
    
\def\MIS{}
\def\mytabcolsep{1.8pt}
\def\myarraystretch{0.8}

\def\minitable#1{\begin{tabular}{c}#1\end{tabular}}
\let\difsize\small
\def\posdif#1{\bgroup\difsize(\textcolor{green}{+#1})\egroup}
\def\nodif#1{\bgroup\difsize(\textcolor{black}{+#1})\egroup}
\def\negdif#1{\bgroup\difsize(\textcolor{black}{-#1})\egroup}

\def\resultsOne{
  \bgroup
  \def\arraystretch{\myarraystretch}
  \setlength{\tabcolsep}{\mytabcolsep}
  \begin{tabular}{lcccccc} \toprule
    Dataset  &         \multicolumn{2}{c}{VOC} & \multicolumn{2}{c}{CCV} & \multicolumn{2}{c}{E-MNIST} \\ \midrule
    Metric            & ACC       & NMI       & ACC       & NMI       & ACC       & NMI       \\ \midrule
    SC(1)             & 38.4      & 39.2      & 10.2      & 0.5       & \MIS      & \MIS      \\
    SC(2)             & 40.2      & 41.1      & 18.8      & 17.3      & \MIS      & \MIS      \\
    SC(3)             & \MIS      & \MIS      & 11.3      & 0.8       & \MIS      & \MIS      \\
    SC(con)           & 37.2      & 38.7      & 9.3       & 7.4      & \MIS      & \MIS      \\ \midrule
    RMKMC             & 45.8      & 46.9      & 17.6      & 16.5      & \MIS      & \MIS      \\
    tRLMvc            & 53.4      & 54.7      & 21.2      & 22.6      & \MIS      & \MIS      \\
    CSMSC             & 48.8      & 49.6      & 19.4      & 18.6      & \MIS      & \MIS      \\
    WMSC              & 47.1      & 46.2      & 20.5      & 19.6      & \MIS      & \MIS      \\
    MCGC              & 52.7      & 54.6      & 22.4      & 21.6      & \MIS      & \MIS      \\
    DCCA              & 39.7      & 42.5      & 17.3      & 18.2      & 47.6      & 44.3      \\
    DMSC              & 54.1      & 53.8      & 18.3      & 19.4      & 65.3      & 61.4      \\
    DAMC              & 56.0      & 55.2      & 24.3      & 23.1      & 64.6      & 59.4      \\
    EAMC              & \BB{60.7} & \BB{61.5} & \BB{26.1} & \BB{26.6} & 66.8      & 62.8      \\ \midrule
    \baseModel & \minitable{55.1\\\negdif{5.6}} & \minitable{\BB{61.5}\\\nodif{0.0}} & \minitable{14.4\\\negdif{11.7}} & \minitable{11.2\\\negdif{15.4}} & \minitable{\BB{86.2}\\\posdif{19.4}} & \minitable{\BB{82.6}\\\posdif{19.8}} \\[0.3cm]
    \contrastiveModel & \minitable{\BB{61.9}\\\posdif{1.2}} & \minitable{\BB{67.5}\\\posdif{6.0}} & \minitable{\BB{29.5}\\\posdif{3.4}} & \minitable{\BB{28.7}\\\posdif{2.1}} & \minitable{\BB{95.5}\\\posdif{28.7}} & \minitable{\BB{90.7}\\\posdif{27.9}} \\
    \bottomrule
  \end{tabular}
  \egroup
}

\def\resultsTwo{
  \bgroup
  \def\arraystretch{\myarraystretch}
  \setlength{\tabcolsep}{\mytabcolsep}
  \begin{tabular}{lcccccc} \toprule
    Dataset  & \multicolumn{2}{c}{E-FMNIST} & \multicolumn{2}{c}{COIL-20} & \multicolumn{2}{c}{RGB-D} \\ \midrule
    Metric              & ACC                                 & NMI                             & ACC                                  & NMI                                  & ACC                                 & NMI                 \\ \midrule
    EAMC                & 55.2                                & \BB{62.5}                       & 69.0                                 & 75.3                                 & 32.3                                & 20.7                \\ \midrule
    \baseModel          & \minitable{\BB{56.8}\\\posdif{1.6}} & \minitable{50.7\\\negdif{11.8}} & \minitable{\BB{77.5}\\\posdif{8.5}}  & \minitable{\BB{91.8}\\\posdif{16.5}} & \minitable{\BB{39.6}\\\posdif{7.3}} & \minitable{\BB{35.6}\\\posdif{14.9}} \\[0.3cm]
    \contrastiveModel   & \minitable{\BB{59.5}\\\posdif{4.3}} & \minitable{52.3\\\negdif{10.2}} & \minitable{\BB{89.4}\\\posdif{20.4}} & \minitable{\BB{95.7}\\\posdif{20.4}} & \minitable{\BB{41.3}\\\posdif{9.0}} & \minitable{\BB{40.5}\\\posdif{19.8}} \\
    \bottomrule
  \end{tabular}
  \egroup
}

    \begin{table}
      \centering
      \resultsOne
      \caption{Clustering metrics [\( \% \)] on VOC, CCV, and E-MNIST. The best and second best are highlighted in bold. The differences between our models and the best baseline model are shown in parentheses. Green differences indicate improvements. Baseline results are taken from \cite{zhouEndtoEndAdversarialAttentionNetwork2020}.}
      \label{tab:mainResults}
    \end{table}

    \begin{table}
      \centering
      \resultsTwo
      \caption{Clustering metrics [\( \% \)] on E-FMNIST, COIL-20 and RGB-D. Same formatting as in Table \ref{tab:mainResults}.}
      \label{tab:mainResults2}
    \end{table}

    \customparagraph{Quantitive results} on VOC, CCV and E-MNIST are shown in Table \ref{tab:mainResults}. The results illustrate that not aligning representations can have a significant improvement (relative gain in ACC larger than 29\% on E-MNIST) compared to adversarial alignment, while selective alignment always improves performance. Note that entries for E-MNIST in Table \ref{tab:mainResults} are missing as the number of samples makes the traditional approaches computationally infeasible.

    Table~\ref{tab:mainResults2} compares \baseModel and \contrastiveModel to the previous state of the art, EAMC on E-FMNIST, COIL-20 and RGB-D. Again, we observe that na{\"i}vely aligning feature representations tends to worsen performance. This highlights the importance of being considerate when aligning representations in multi-view clustering.

  \customparagraph{Ablation study.}
    \begin{table}
      \centering

\bgroup
\def\TRUE{\ding{51}}
\def\FALSE{--}
\def\arraystretch{0.85}
\def\DATASET#1{\multirow{4}{*}{\rotatebox{90}{#1}}}

\begin{tabular}{ccccc} \toprule
                         & Neg.\ samp.\ & Ad.\ weight & ACC \( [\%] \) & NMI \( [\%] \) \\ \midrule
 \DATASET{E-MNIST}       & \FALSE       & \FALSE      & 87.4           & 86.8 \\
                         & \FALSE       & \TRUE       & 94.7           & 89.5 \\
                         & \TRUE        & \FALSE      & 87.5           & 86.6 \\
                         & \TRUE        & \TRUE       & \BB{95.5}      & \BB{90.7} \\ \midrule
  \DATASET{VOC}          & \FALSE       & \FALSE      & 54.7           & 61.3 \\
                         & \FALSE       & \TRUE       & 55.3           & 60.7 \\
                         & \TRUE        & \FALSE      & 58.5           & 67.4 \\
                         & \TRUE        & \TRUE       & \BB{61.9}      & \BB{67.5} \\
  \bottomrule
\end{tabular}

\egroup

      \caption{Ablation study results for \contrastiveModel on E-MNIST and VOC.
      }
      \label{tab:ablation}
    \end{table}
    We perform an ablation study in order to evaluate the effects of the different components in the contrastive loss\footnote{We include an ablation study with the DDC loss in the supplementary.}. Specifically, we train \contrastiveModel with and without the proposed negative pair sampling and the adaptive weight factor (\( \min\{w_1, \dots, w_V\} \)), on E-MNIST and VOC. When we remove the negative sampling, we construct \( \Neg(z_i^{(v)}, z_i^{(u)}) \) by including the similarities between all views of object \( i \), and all views of all the other objects within the current batch.

    Results of the ablation study (Table \ref{tab:ablation}) show that dropping the adaptive weighting and the negative sampling strategy both have a negative impact on \contrastiveModel's performance. This justifies their inclusion in the final contrastive loss.

  \customparagraph{View prioritization.}
    \begin{table}
      \centering

\bgroup
\def\arraystretch{0.85}
\setlength{\tabcolsep}{4.5pt}
\begin{tabular}{lccccccccc} \toprule
           & \multicolumn{3}{c}{EAMC} & \multicolumn{3}{c}{\baseModel} & \multicolumn{3}{c}{\contrastiveModel}  \\ \midrule
  View     & 1  & 2  & 3  & 1  & 2  & 3  & 1  & 2  & 3  \\ \midrule
  VOC      & 48 & 52 &    & 47 & 53 &    & 64 & 36 &    \\
  CCV      & 26 & 38 & 36 & 32 & 35 & 33 & 1 & 75 & 24 \\
  E-MNIST  & 48 & 52 &    & 95 & 05 &    & 67 & 33 &    \\
  E-FMNIST & 53 & 47 &    & 78 & 22 &    & 99 & 1 &    \\
  COIL-20  & 32 & 32 & 36 & 33 & 35 & 32 & 34 & 32 & 34 \\
  RGB-D    & 53 & 47 &    & 59 & 41 &    & 59 & 41 &    \\
  \bottomrule
\end{tabular}
\egroup

      \caption{Fusion weights [\%] for EAMC, \baseModel, and \contrastiveModel. For EAMC, we split the entire dataset into batches of size \( 100 \) and report the average weight over these batches.}
      \label{tab:weights}
    \end{table}
    Table~\ref{tab:weights} shows the weight parameters that are obtained for EAMC, \baseModel and \contrastiveModel for all datasets. EAMC always produces close to uniform weight distributions, while \baseModel and \contrastiveModel are able to suppress uninformative views. Note, for datasets, such as COIL-20, where views are assumed equally important\footnote{Since views in COIL-20 refer to objects depicted from random angles.}, we do also observe close to uniform weight distributions for \baseModel and \contrastiveModel.

    To further assess our models' capabilities to prioritize views, we corrupt the edge-view (view \( 2 \)) in E-MNIST with additive Gaussian noise, and record the models' performance as the standard deviation of the noise increases. We also repeat the experiment for the EAMC model, as it represents the current state of the art.
    \begin{figure}
      \centering
      \begin{subfigure}{0.49\columnwidth}
        \includegraphics[width=\textwidth]{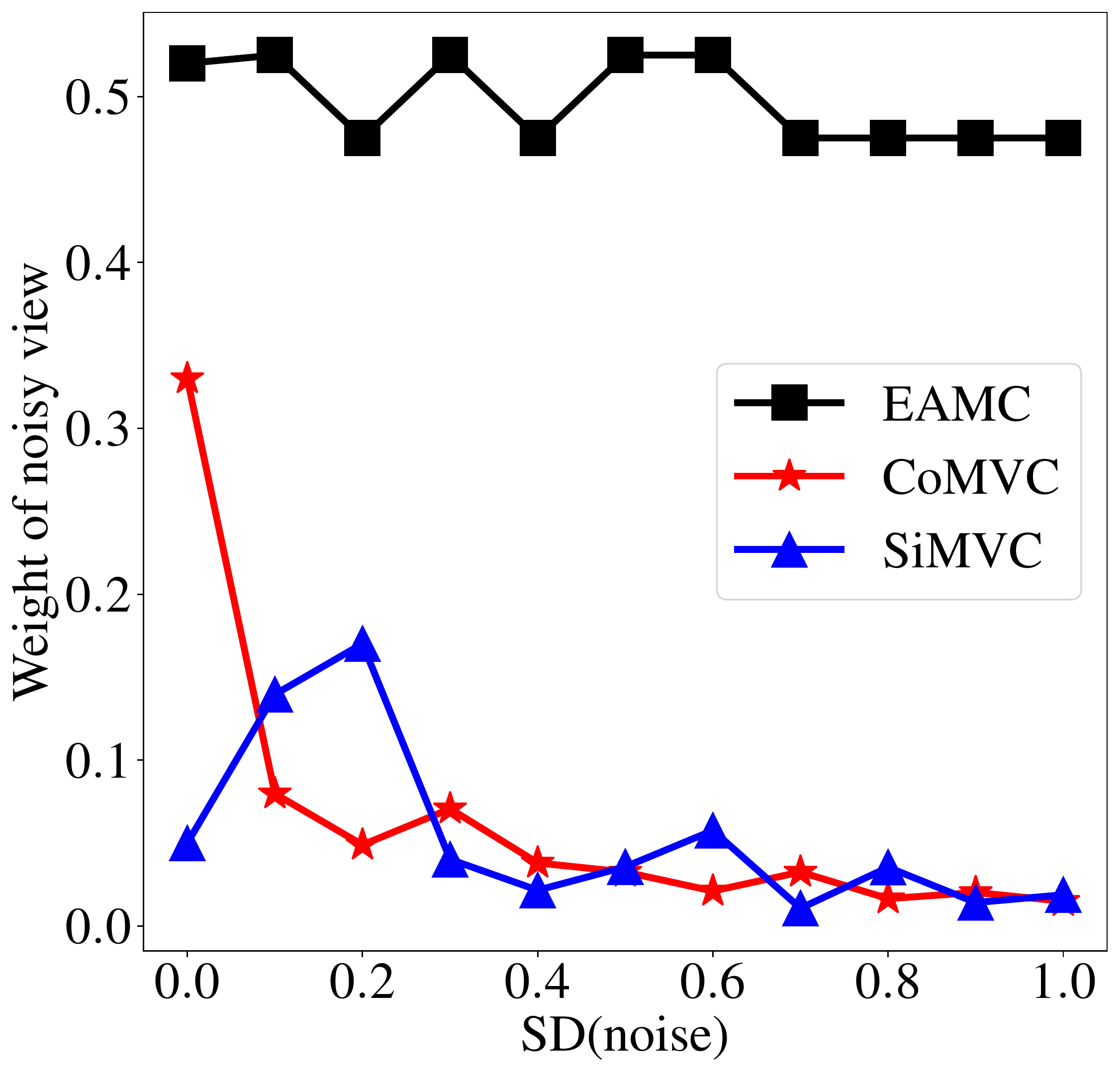}
      \end{subfigure}
      \begin{subfigure}{0.49\columnwidth}
        \includegraphics[width=\textwidth]{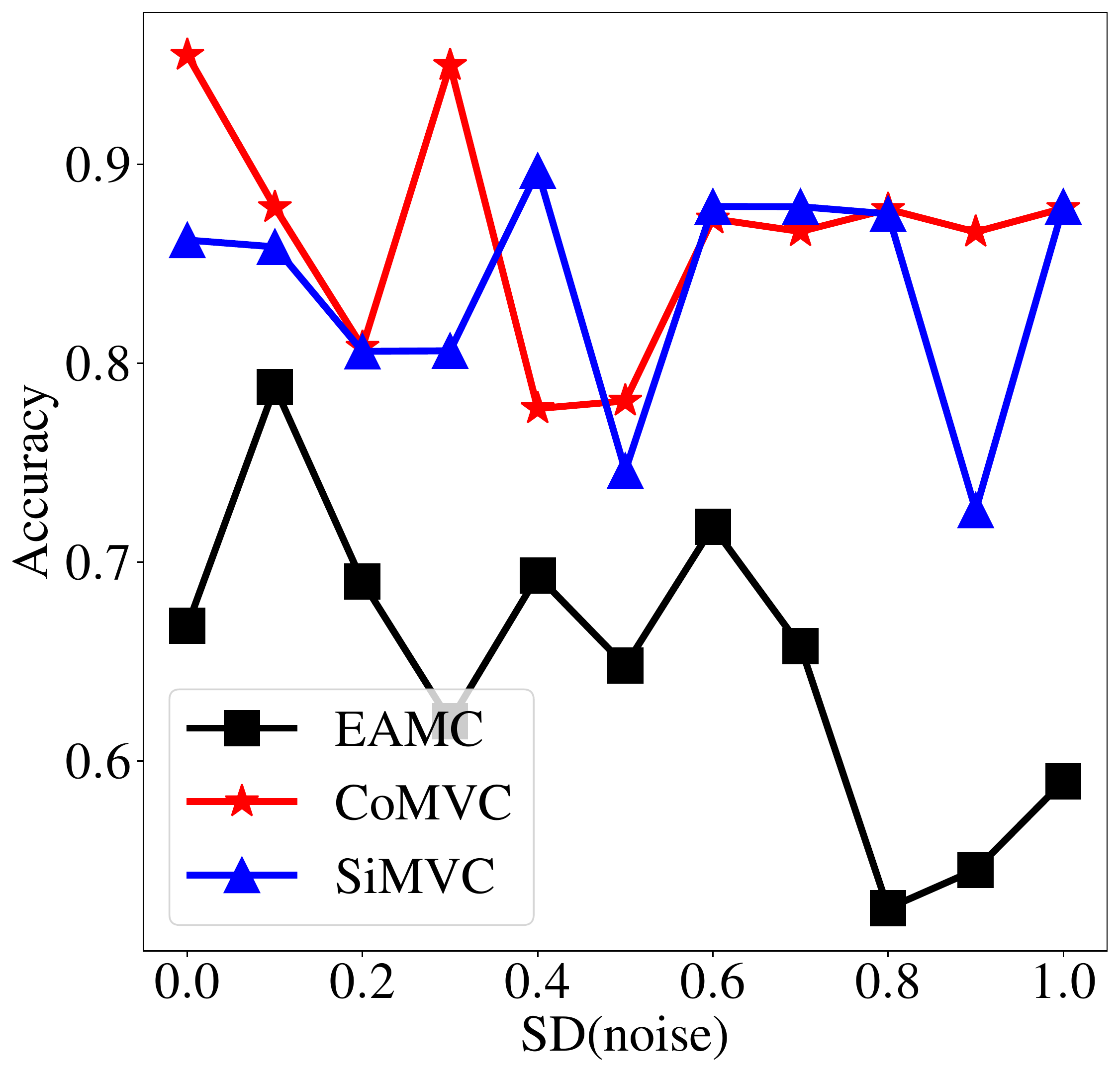}
      \end{subfigure}
      \caption{Fusion weights and clustering accuracy (ACC) on E-MNIST, with increasing levels of Gaussian noise added to the second view.
      }
      \label{fig:noiseExperiment}
    \end{figure}
    Figure \ref{fig:noiseExperiment} shows the resulting fusion weights for the noisy view and the clustering accuracies, for different noise levels. For \baseModel and \contrastiveModel, we observe that the weight of the noisy view decreases as the noise increases. The mechanism for prioritizing views thus works as expected. \baseModel and \contrastiveModel can therefore produce accurate clusterings, regardless of the noise level. Conversely, we observe that the attention mechanism in EAMC is unable to produce fusion weights that suppress the noisy view. This results in a significant drop in clustering accuracy, as the noise increases.

  \customparagraph{Selective alignment in \contrastiveModel.}
    \begin{figure}
      \centering
      \begin{subfigure}{0.9\columnwidth}
        \includegraphics[width=0.49\textwidth]{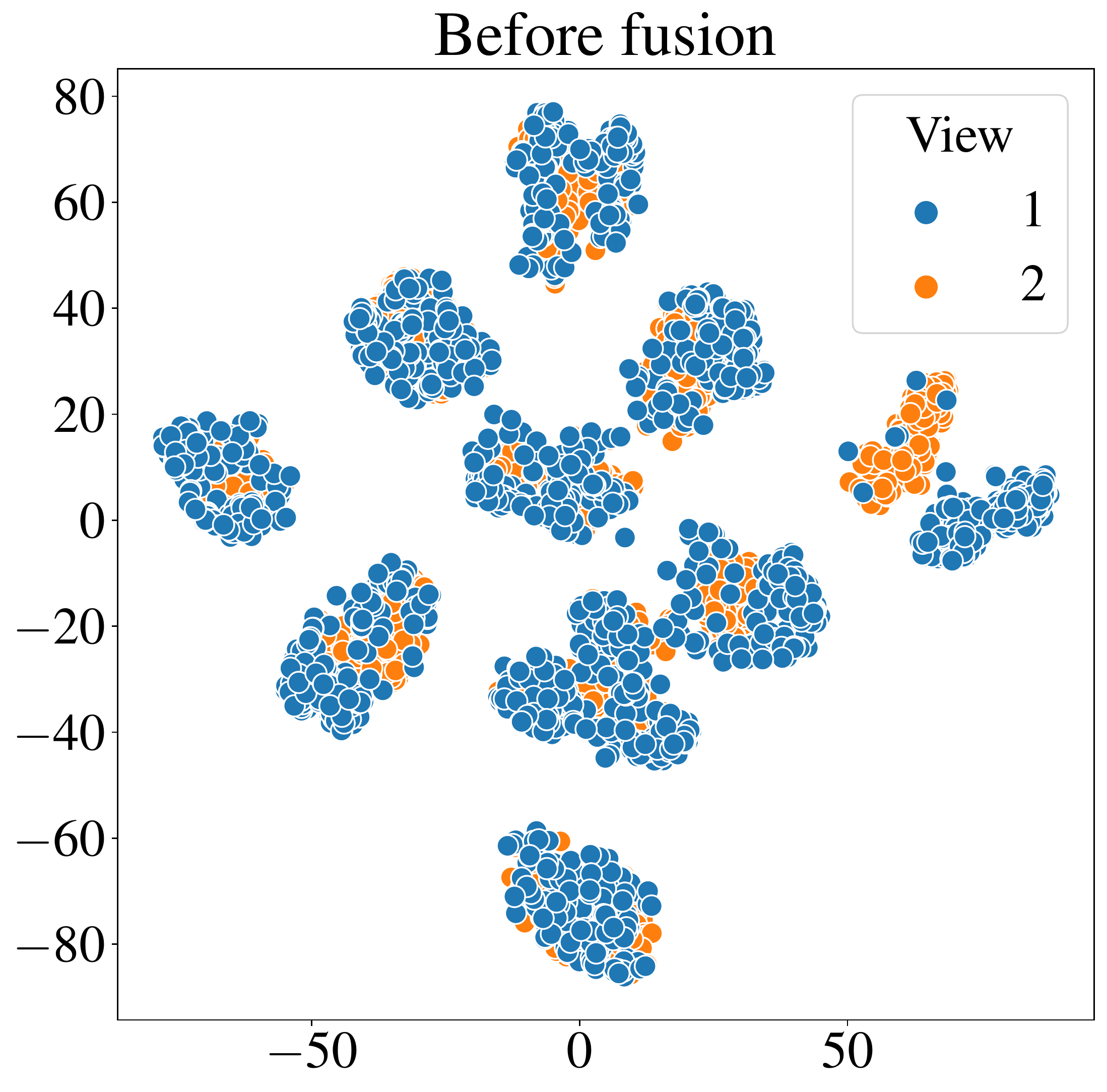}
        \includegraphics[width=0.49\textwidth]{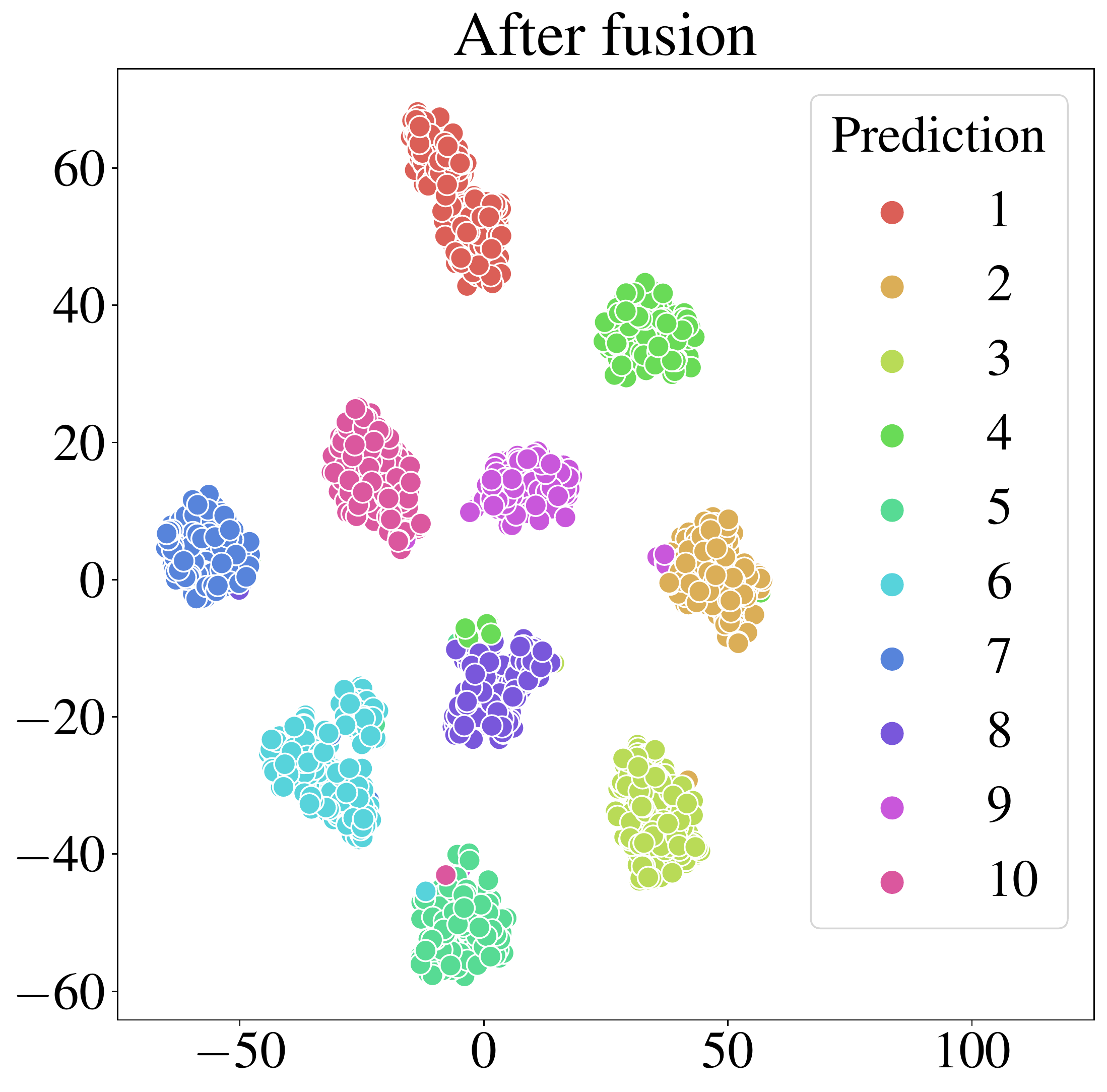}
      \end{subfigure}
      \begin{subfigure}{0.9\columnwidth}
        \includegraphics[width=0.49\textwidth]{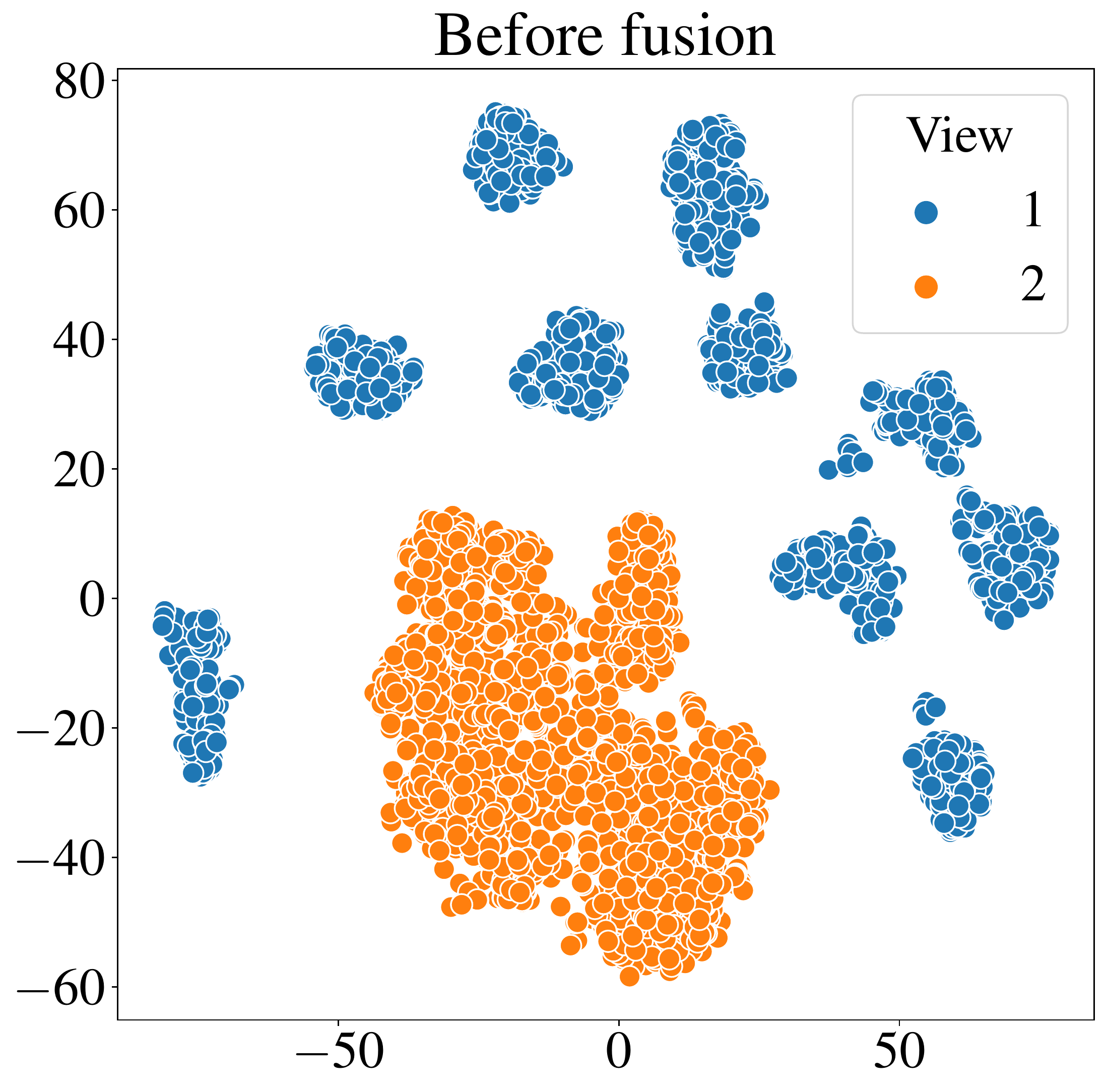}
        \includegraphics[width=0.49\textwidth]{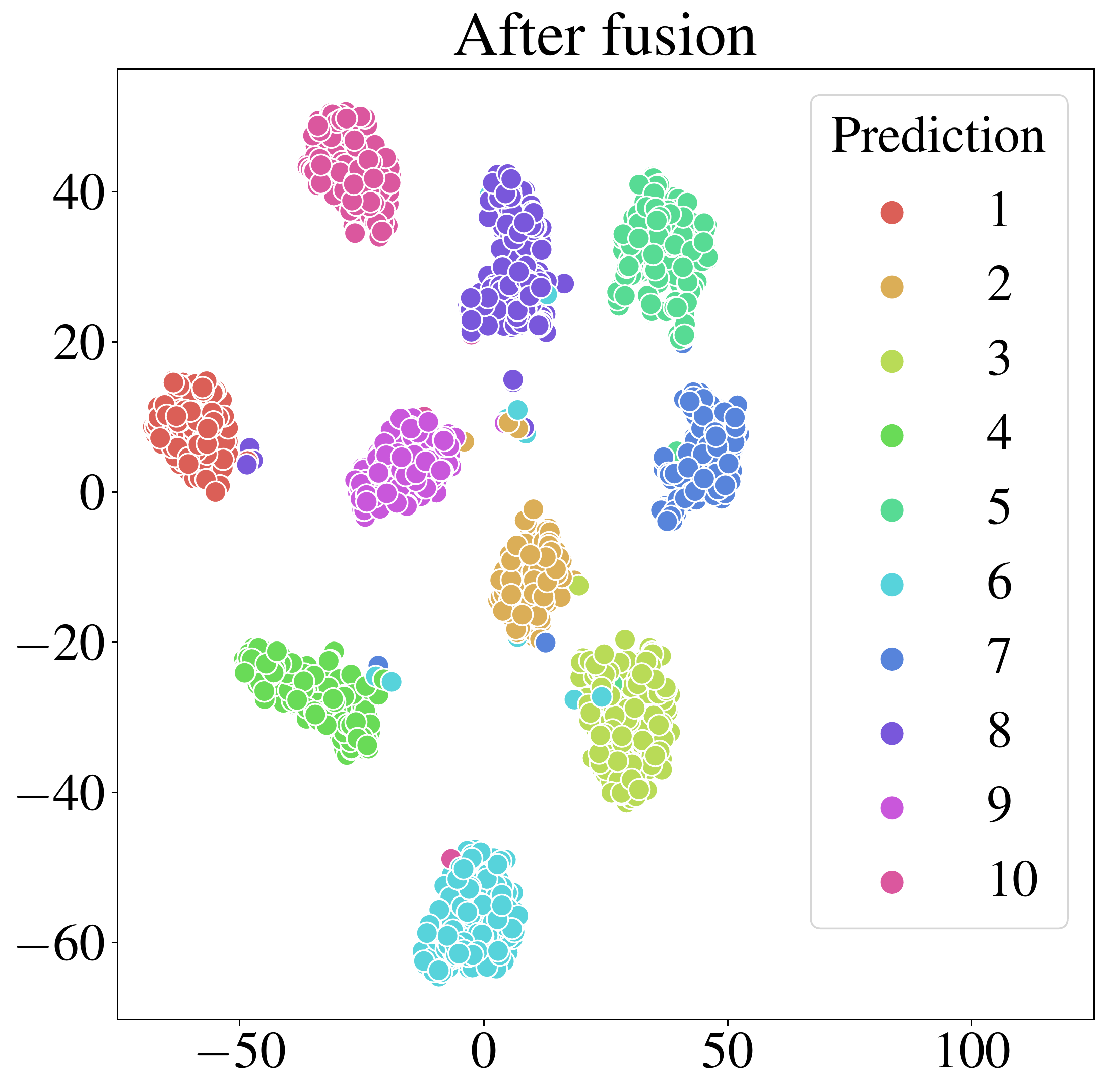}
      \end{subfigure}
      \caption{Learned representations before and after fusion for regular (top) and noisy (\( \sigma = 1 \)) E-MNIST (bottom). Projected to 2-D using T-SNE.
      }
      \label{fig:mnistAligned}
    \end{figure}
    Figure \ref{fig:mnistAligned} demonstrates the selective alignment in \contrastiveModel, for the noise-free and noisy variants of the E-MNIST dataset. In the noise-free case, \contrastiveModel aligns the representations, resulting in clusters that are well separated. When the second view has been corrupted by noise however, it is discarded by the view prioritization mechanism, by setting its fusion weight to \( 0 \). This simultaneously disables the alignment procedure, preventing the fused representation from being corrupted by the noisy view, thus preserving the cluster structure.

\section{Conclusion}
  \label{sec:conclusion}
  Our work highlights the importance of considering representation alignment when performing multi-view clustering. Comparing the results of our \baseModel to previous results illustrates that na{\"i}vely aligning distributions using adversarial learning can prevent the model from learning good clusterings, while \contrastiveModel illustrates the benefit of selective alignment, leveraging the best of both worlds.

\vspace{0.5em}
\customparagraph{Acknowledgements.}
This work was financially supported by the Research Council of Norway (RCN), through its Centre for Research-based Innovation funding scheme (Visual Intelligence, grant no. 309439), and Consortium Partners. The work was further funded by RCN FRIPRO grant no. 315029, RCN IKTPLUSS grant no. 303514, and the UiT Thematic Initiative ``Data-Driven Health Technology''

{\small
\bibliographystyle{ieee_fullname}
\bibliography{bibliography}
}

  \clearpage
  \appendix
  {\begin{center} \Large \bfseries Supplementary material \end{center}}

  \section{Pitfalls of distribution alignment in multi-view clustering}
    \subsection{Proof sketch for Proposition 1}
      
\begin{proof}[Proof sketch]
  Suppose that, for view \( v \), the \( k_v \) clusters in the input space, are mapped to \( c_v \) unique points in the representation space. This is possible under assumptions 1 and 2. The number of unique clusters after fusion is then upper bounded by the number of unique linear combinations on the form:
  \begin{align}
    \sum\limits_{v=1}^{V} w_v c_\star^{(v)}
  \end{align}
  where \( c_\star^{(v)} \) is one of the \( c_v \) points obtained by mapping the \( k_v \) unique points (clusters) for view \( v \), to the representation space. Note that the encoders might not be injective, meaning that we can have \( c_v < k_v \). Under assumption 3, the maximum number of unique such linear combinations is equal to \( c_1 \cdot c_2 \cdots c_V = \prod_{v=1}^{V} c_v \). Since we only have \( k \) clusters in the entire dataset, the number of unique clusters after fusion will also be upper bounded by \( k \). This gives:
  \begin{align}
      \kappa^\text{fused}_\cdot = \min\lrc{k, \prod_{v=1}^{V} c_v}.
  \end{align}

  \customparagraph{Perfectly aligned representations.}
  Clusters that are separated in the input space can be mapped to the same centroid in the representation space, but not vice versa. I.e. it is not possible for the encoding network to separate two clusters that lie at the same point in the input space. The perfect alignment constraint therefore forces the number of unique points to be equal to the smallest \( k_v \) for each cluster. That is:
  \begin{align}
    c_v = \min\limits_{w=1, \dots, V}\{k_w\}, \quad v = 1, \dots, V.
  \end{align}
  We then get
  \begin{align}
    \kappa^\text{fused}_\text{aligned} &= \min\lrc{k, \prod\limits_{v=1}^{V} \min\limits_{w=1, \dots, V}\{k_w\}} \\
    &= \min\lrc{k, \left(\min\limits_{v=1, \dots, V}\{k_v\} \right)^V}
  \end{align}

  \customparagraph{Unaligned representations.}
  Here the encoder for view \( v \) has the ability to map the \( k_v \) separable clusters to \( k_v \) unique representations, which do not coincide with the representations from any other views. We therefore get \( c_v = k_v \), and
  \begin{align}
    \kappa^\text{fused}_\text{not aligned} &= \min\lrc{k, \prod\limits_{v=1}^{V} c_v} \\
    &= \min\lrc{k, \prod\limits_{v=1}^{V} k_v}.
  \end{align}
\end{proof}

    \subsection{Experiments with toy data}
      
\begin{figure*}
  \centering
  \begin{subfigure}[t]{0.245\textwidth}
    \centering
    \includegraphics[height=3.8cm]{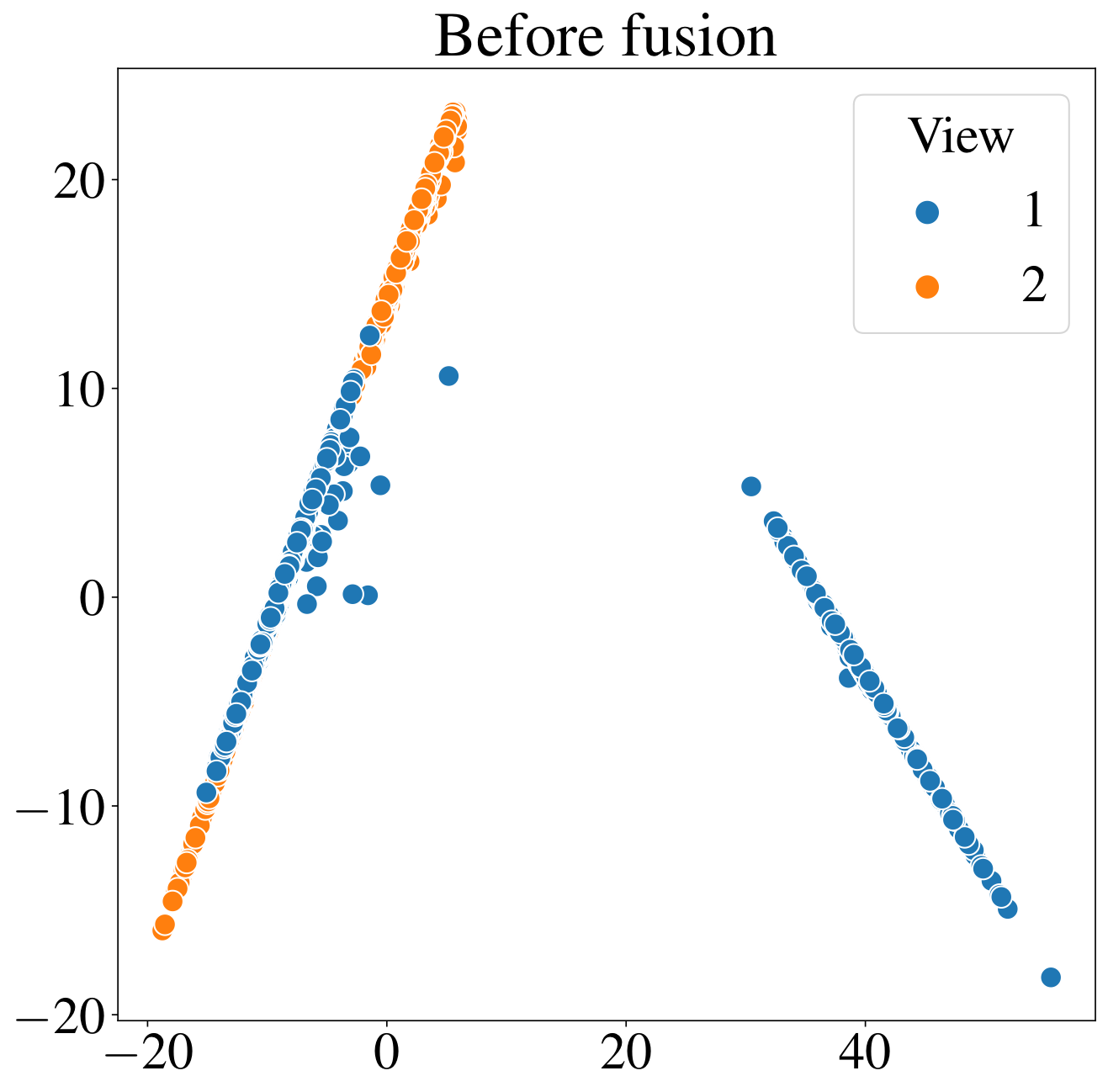}
    \includegraphics[height=3.8cm]{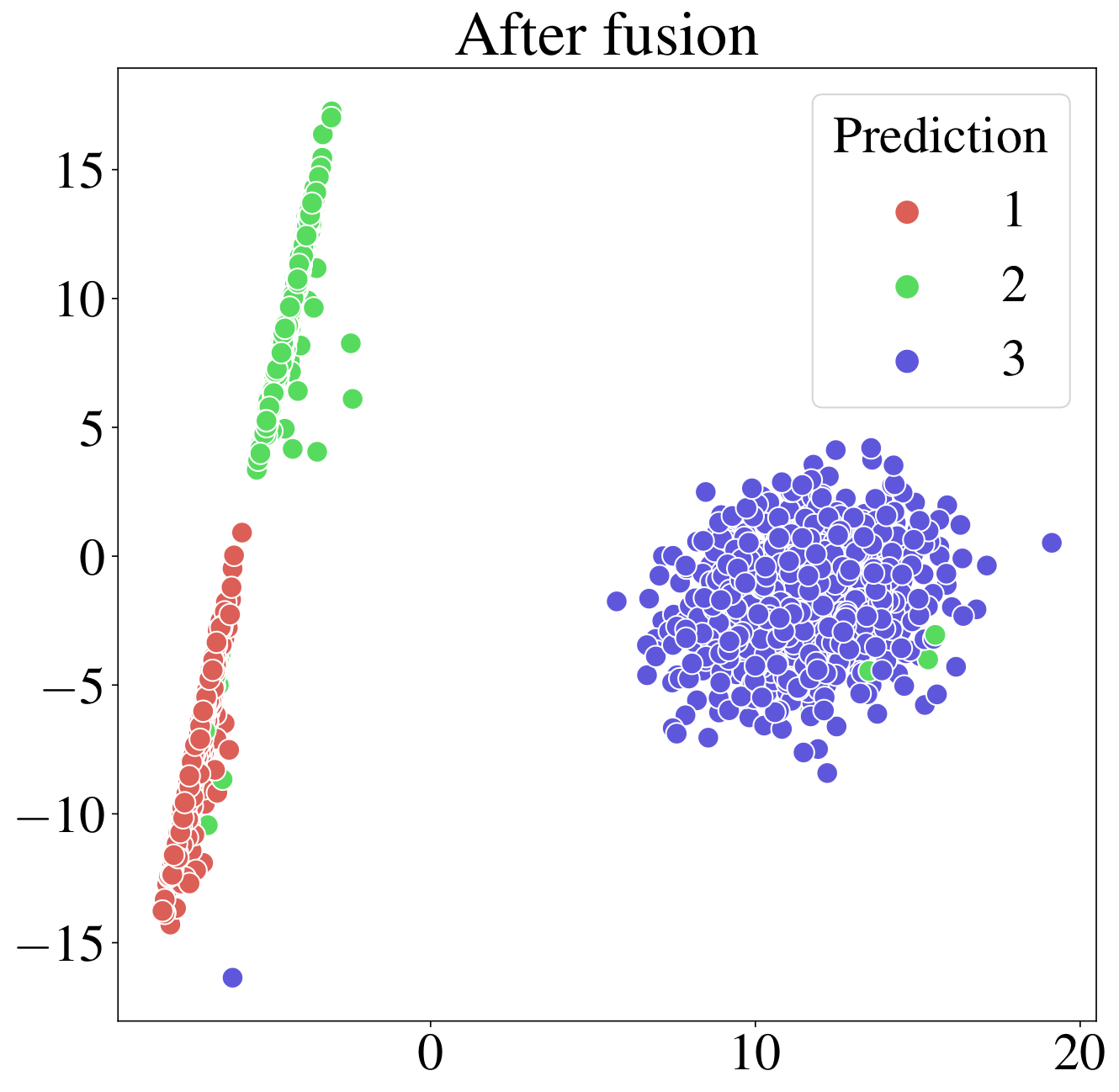}
    \caption{\baseModel+Adv. ACC \( = 0.99 \)}
    \label{fig:toyDataResults_base_adv_supp}
  \end{subfigure}
  \begin{subfigure}[t]{0.245\textwidth}
    \centering
    \includegraphics[height=3.8cm]{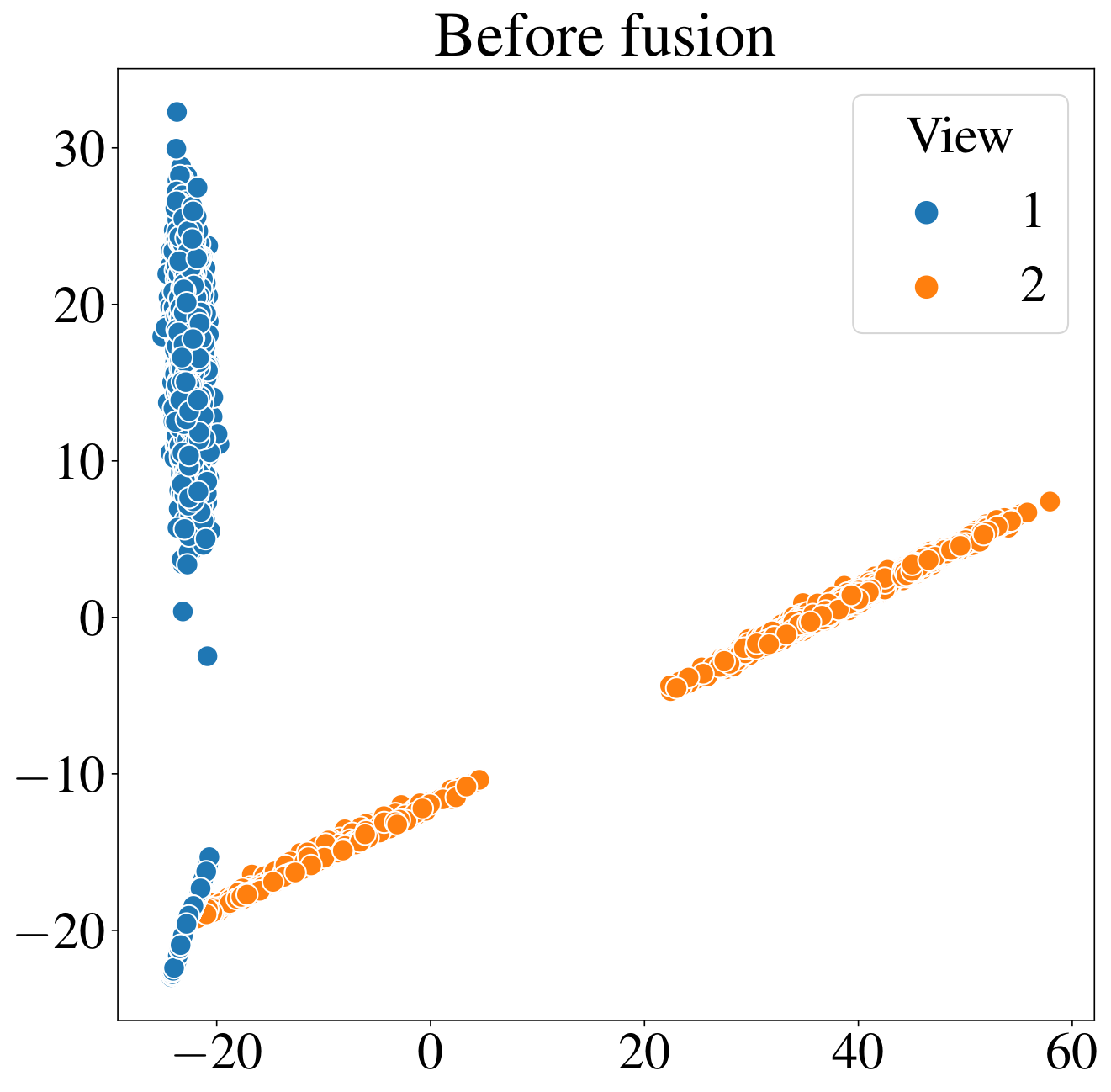}
    \includegraphics[height=3.8cm]{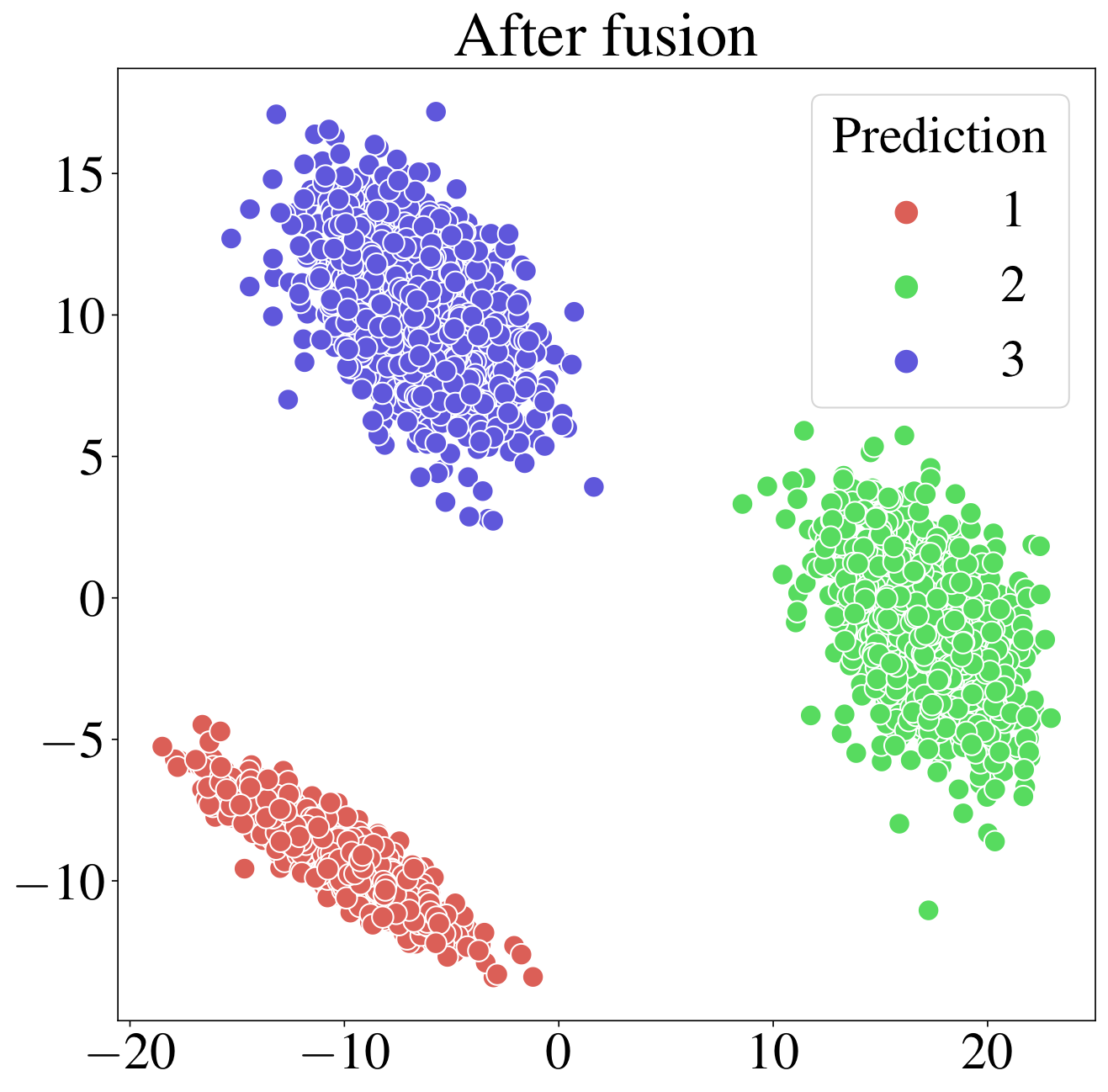}
    \caption{\baseModel. ACC \( = 1.0 \).}
    \label{fig:fig:toyDataResults_base_supp}
  \end{subfigure}
  \begin{subfigure}[t]{0.245\textwidth}
    \centering
    \includegraphics[height=3.8cm]{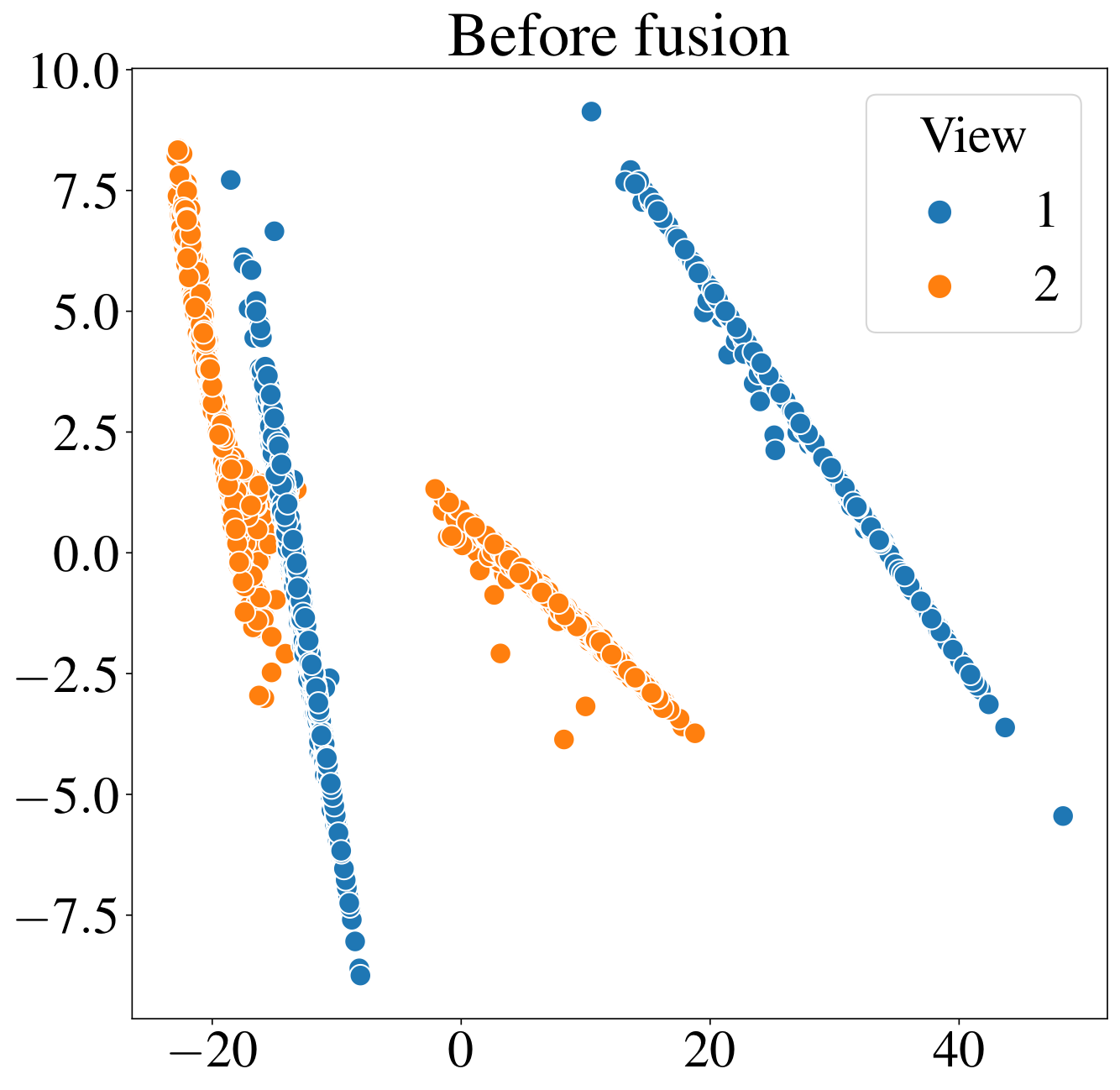}
    \includegraphics[height=3.8cm]{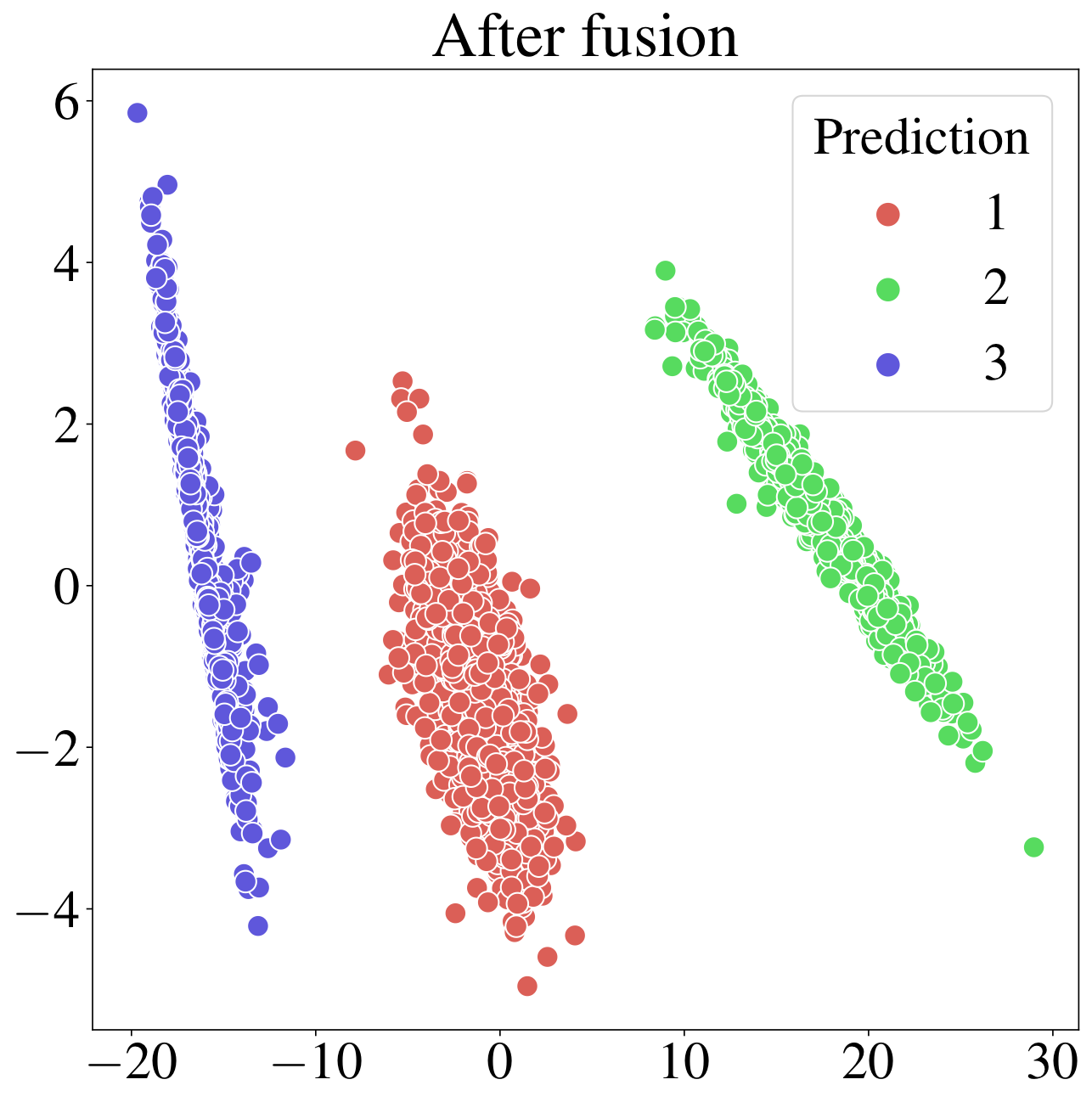}
    \caption{\contrastiveModel. ACC = \( 1.0 \).}
    \label{fig:fig:toyDataResults_contrastive_supp}
  \end{subfigure}
  \begin{subfigure}[t]{0.245\textwidth}
    \centering
    \includegraphics[height=3.8cm]{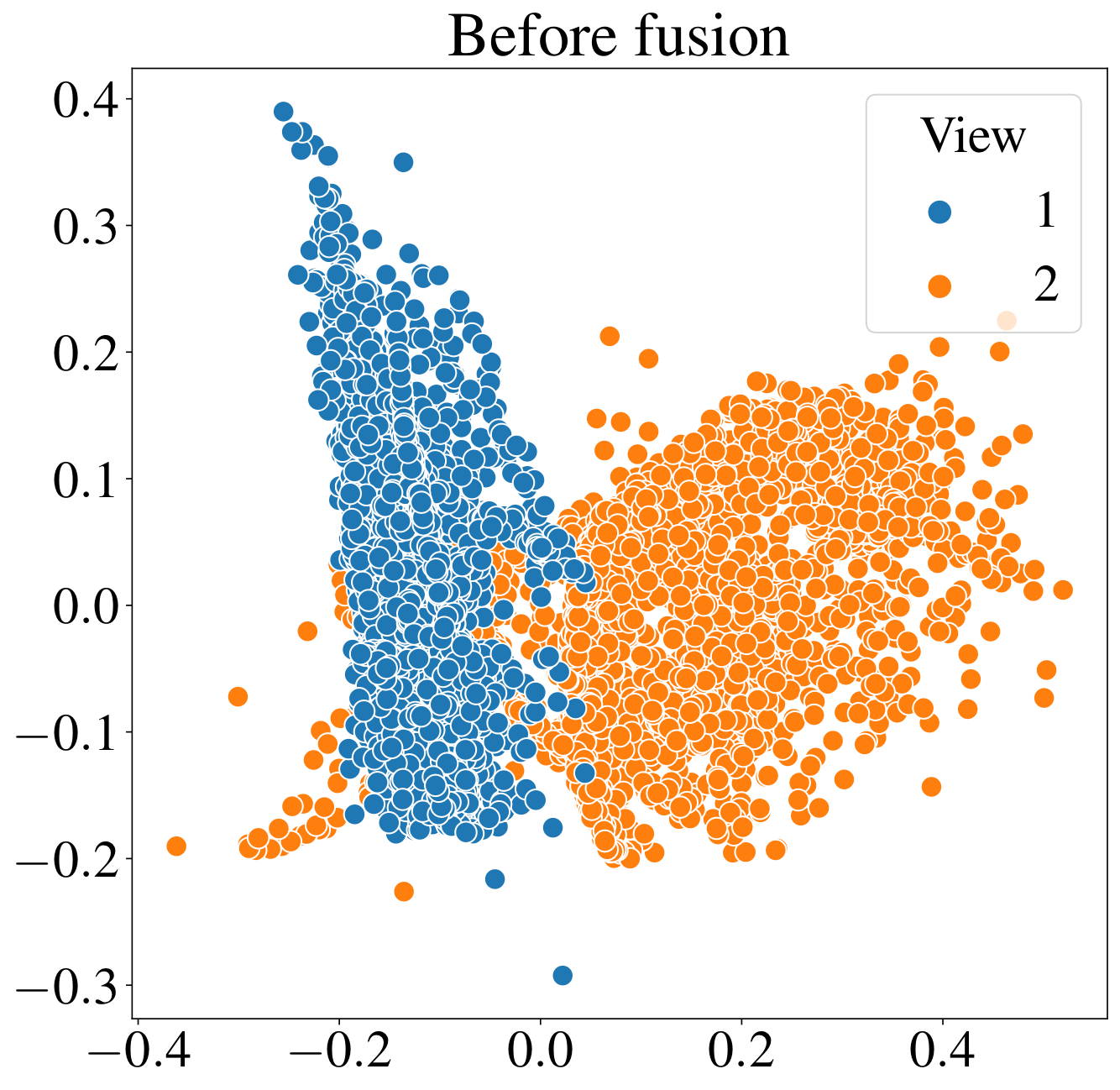}
    \includegraphics[height=3.8cm]{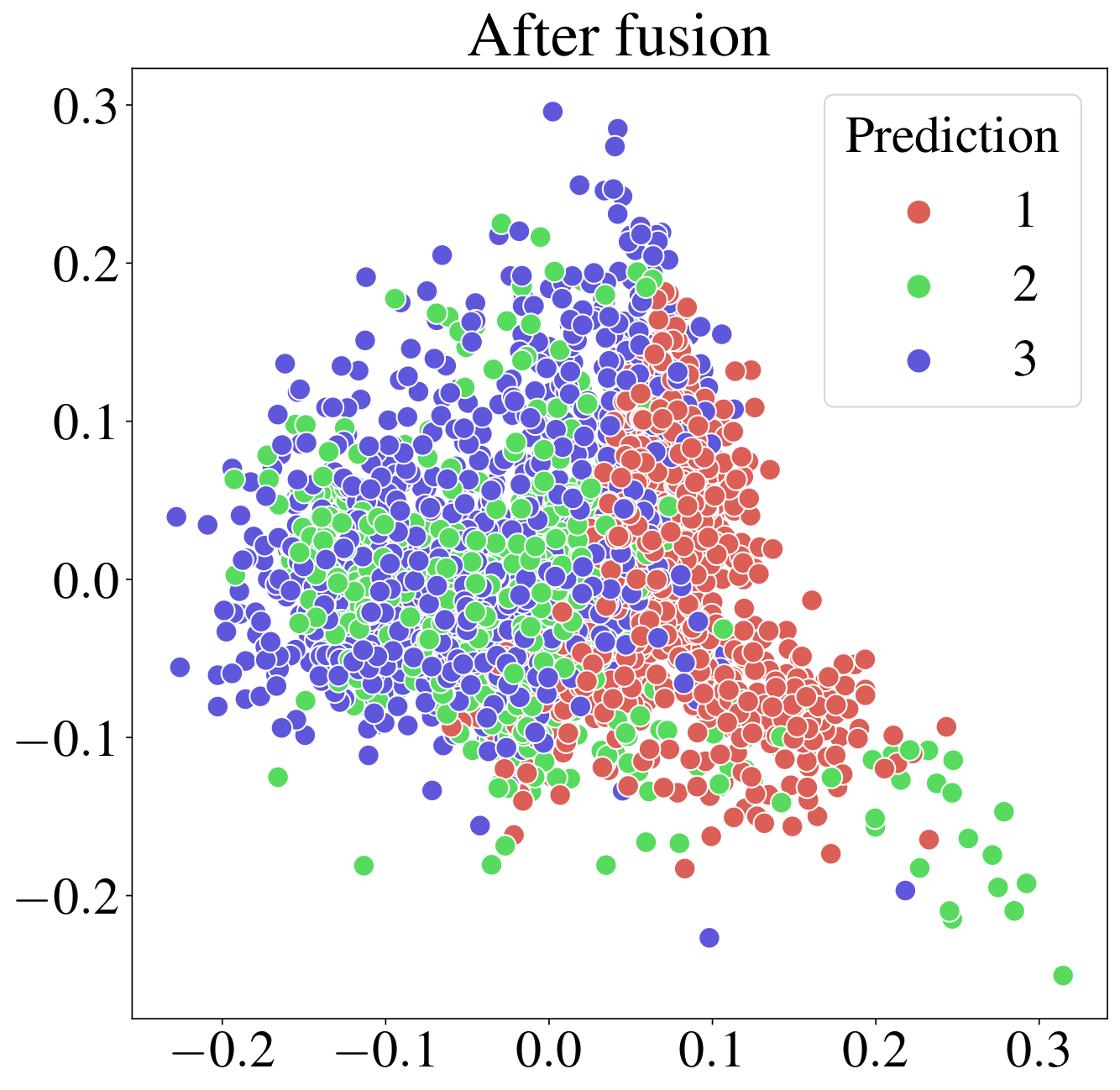}
    \caption{EAMC. ACC \( = 0.44 \) .}
    \label{fig:fig:toyDataResults_eamc_supp}
  \end{subfigure}
  \caption{Representations for \baseModel with and without adversarial alignment, \contrastiveModel, and EAMC on a version of our toy dataset with \( 3 \) clusters.}
  \label{fig:toyDataResults3}
\end{figure*}

\begin{figure}
  \centering
  \includegraphics[height=3.9cm]{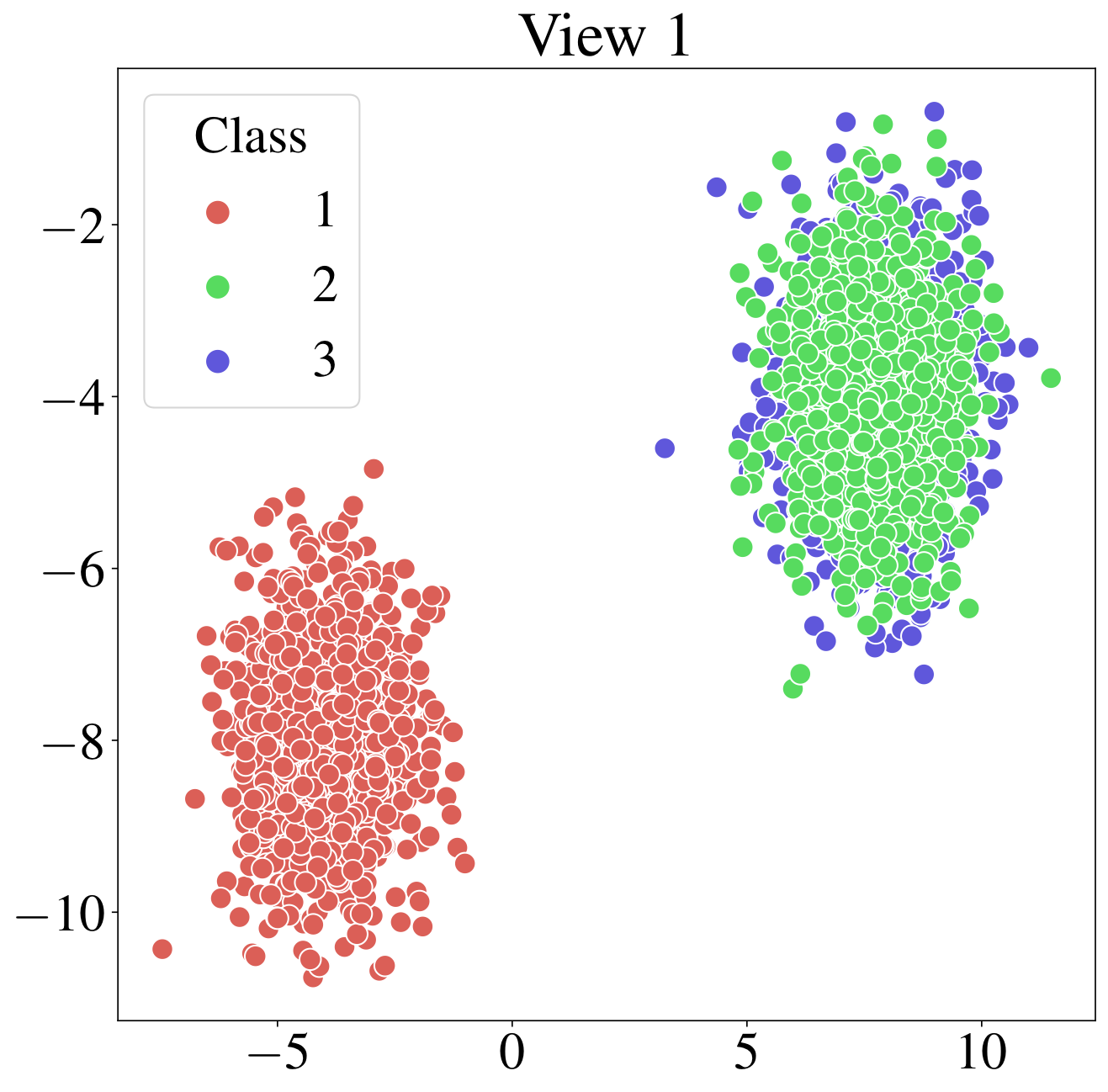}
  \includegraphics[height=3.9cm]{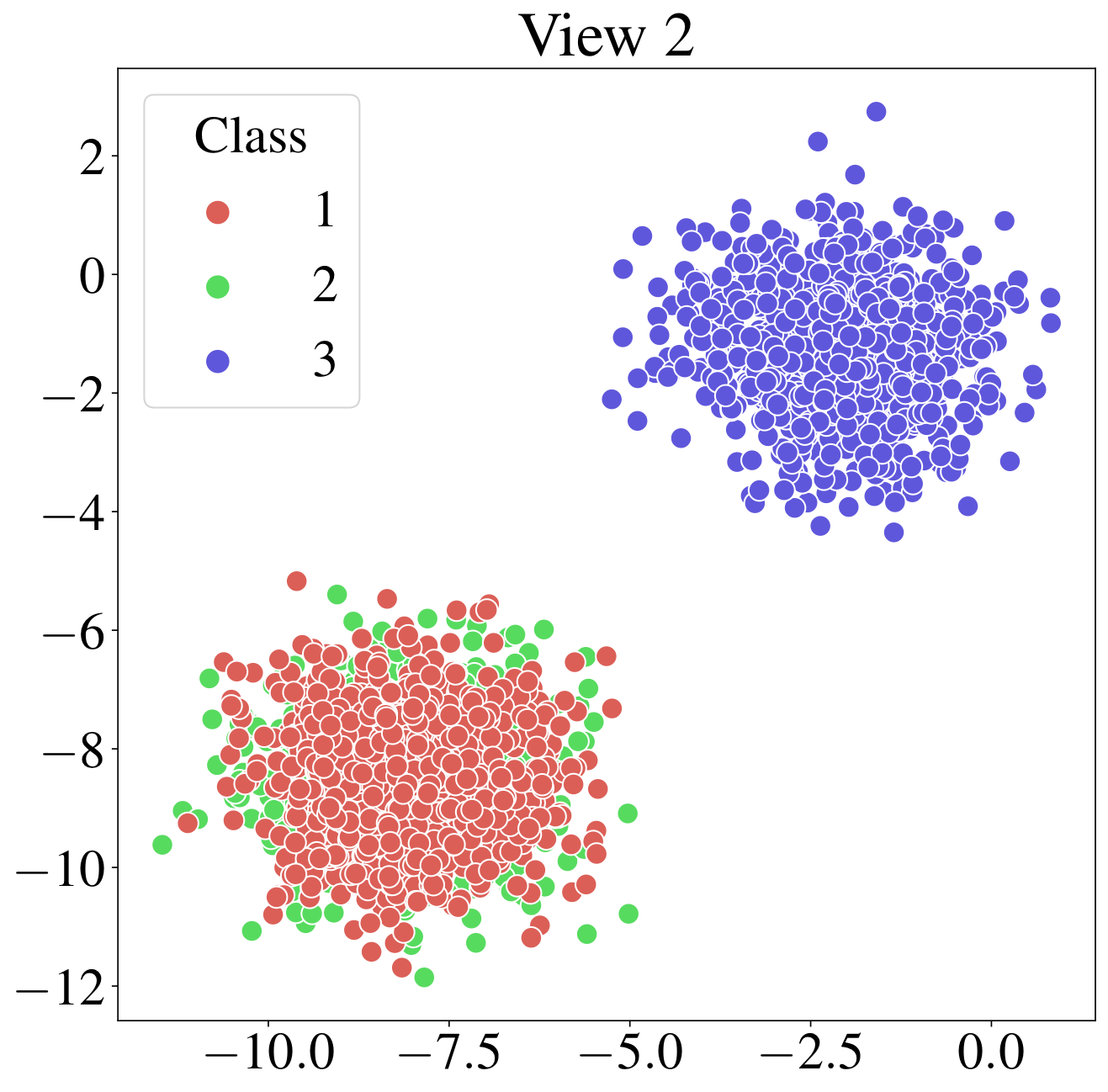}
  \caption{Toy dataset with 3 clusters. View 1: Class 1 is isolated, and classes (2,3) overlap. View 2: Class (1,2) overlap, and class 3 is isolated.}
  \label{fig:toydataset3}
\end{figure}

Figure \ref{fig:toyDataResults3} shows the results of our toy experiment with \( 3 \) clusters instead of \( 5 \). The dataset is shown in Figure \ref{fig:toydataset3}. Similarly to the experiment with \( 5 \) clusters, we observe that \baseModel + Adv.\ partially aligns the distributions. Due to the reduced number of clusters however, it is still possible to separate the clusters after fusion. This outcome is consistent with Proposition 1, from which we get \( \kappa^\text{fused}_\text{aligned} = \min\{3, 2^2\} = 3 \).

For \contrastiveModel, we see that the angles between representations have been aligned, which also results in separable clusters.

EAMC attempts to align the distributions, which results in all clusters being mixed together after fusion -- similar to what we observed for the experiment with \( 5 \) clusters. For this experiment, we also observe that EAMC produces approximately equal fusion weights. This violates assumption 3 in Proposition 1, and can further reduce the cluster separability in the space of fused representations.

  \section{Methods}
    \subsection{\contrastiveModel with projection head}

\begin{table}
  \bgroup
  \setlength{\tabcolsep}{2.4pt}
  \def\DATASET#1{\multirow{8}{*}{\rotatebox{90}{#1}}}
  \centering
  \begin{tabular}{cccccc}\toprule
    Dataset           & \smalltable{Projection\\head} & \smalltable{Negative\\sampling} & \smalltable{Adaptive\\weight} & \smalltable{ACC\\\( [\%] \)} & \smalltable{NMI\\\( [\%] \)} \\ \midrule
    \DATASET{E-MNIST} & \MIS                          & \MIS                            & \MIS                          & 87.4           & 86.8 \\
                      & \MIS                          & \TRUE                           & \MIS                          & 87.5           & 86.6 \\
                      & \MIS                          & \MIS                            & \TRUE                         & 94.7           & 89.5 \\
                      & \MIS                          & \TRUE                           & \TRUE                         & \BB{95.5}      & \BB{90.7} \\ \cmidrule{2-6}
                      & \TRUE                         & \MIS                            & \MIS                          & 87.5           & 86.9 \\
                      & \TRUE                         & \TRUE                           & \MIS                          & 88.2           & 86.3 \\
                      & \TRUE                         & \MIS                            & \TRUE                         & 87.4           & 87.2 \\
                      & \TRUE                         & \TRUE                           & \TRUE                         & 77.1           & 77.5 \\ \midrule
    \DATASET{VOC}     & \MIS                          & \MIS                            & \MIS                          & 54.7           & 61.3 \\
                      & \MIS                          & \TRUE                           & \MIS                          & 58.5           & 67.4 \\
                      & \MIS                          & \MIS                            & \TRUE                         & 55.3           & 60.7 \\
                      & \MIS                          & \TRUE                           & \TRUE                         & 61.9           & \BB{67.5} \\ \cmidrule{2-6}
                      & \TRUE                         & \MIS                            & \MIS                          & 53.4           & 58.2 \\
                      & \TRUE                         & \TRUE                           & \MIS                          & 57.0           & 63.2 \\
                      & \TRUE                         & \MIS                            & \TRUE                         & \BB{62.4}      & 65.3 \\
                      & \TRUE                         & \TRUE                           & \TRUE                         & 55.2           & 59.6 \\
    \bottomrule
  \end{tabular}
  \egroup
  \caption{\contrastiveModel ablation study with and without a projection head.}
  \label{tab:ablation_proj}
\end{table}

Table \ref{tab:ablation_proj} shows the results of an extension of the ablation study for \contrastiveModel, where we also include a projection head between the view representations and the cosine similarity. Following \cite{chenSimpleFrameworkContrastive2020a}, we let the projection head be two fully connected layers, separated by a \( \relu \)-nonlinearity. Batch normalization is applied after both layers. The results show that some configurations benefit marginally from the addition of a projection head. However, adding the projection head does not improve the overall performance of \contrastiveModel. We therefore chose to not include it in the final model.

    \subsection{Ablation study: clustering loss}

\begin{table}
  \centering
  \def\MODEL#1{\multirow{7}{*}{\rotatebox{0}{#1}}}
  \begin{tabular}{ccccccc} \toprule
    Model                     & \( \cl L_1 \) & \( \cl L_2 \) & \( \cl L_3 \) & ACC \( [\%] \) & NMI \( [\%] \) \\ \midrule
    \MODEL{\baseModel}        & \TRUE         & \MIS          & \MIS          & 19.2           & 19.6 \\
                              & \MIS          & \TRUE         & \MIS          & 38.1           & 31.4 \\
                              & \MIS          & \MIS          & \TRUE         & 75.2           & 73.9 \\
                              & \TRUE         & \TRUE         & \MIS          & 78.2           & 78.6 \\
                              & \TRUE         & \MIS          & \TRUE         & 76.6           & 77.5 \\
                              & \MIS          & \TRUE         & \TRUE         & 77.4           & 76.9 \\
                              & \TRUE         & \TRUE         & \TRUE         & \BB{86.2}      & \BB{82.6} \\ \midrule
    \MODEL{\contrastiveModel} & \TRUE         & \MIS          & \MIS          & 19.3           & 20.6 \\
                              & \MIS          & \TRUE         & \MIS          & 36.5           & 25.2 \\
                              & \MIS          & \MIS          & \TRUE         & 72.8           & 71.7 \\
                              & \TRUE         & \TRUE         & \MIS          & 71.3           & 73.2 \\
                              & \TRUE         & \MIS          & \TRUE         & 78.0           & 78.2 \\
                              & \MIS          & \TRUE         & \TRUE         & 74.8           & 73.5 \\
                              & \TRUE         & \TRUE         & \TRUE         & \BB{95.5}      & \BB{90.7} \\ \bottomrule
  \end{tabular}
  \caption{Results of an ablation study where we systematically drop terms from the clustering loss. The checkmarks indicate which terms that are included in each configuration.}
  \label{tab:ddcAblation}
\end{table}

Here, we perform an ablation study on the E-MNIST dataset, in order to show the effects of the individual terms in the DDC \cite{kampffmeyerDeepDivergencebasedApproach2019} clustering loss. Note that, since not all configurations include the \( \cl L_1 \) term, we select models based on the sum of the included terms instead. The resulting accuracies for \baseModel and \contrastiveModel when we systematically drop terms from the clustering loss, are listed in Table \ref{tab:ddcAblation}. These results are in line with previous ablation studies conducted on the DDC clustering loss
\cite{kampffmeyerDeepDivergencebasedApproach2019,zhouEndtoEndAdversarialAttentionNetwork2020}:
The models perform best when all terms are included -- dropping terms from the clustering loss reduces the performance of both \baseModel and \contrastiveModel.

  \section{Experiments}
    \subsection{Source code}
      The source code for our experiments is publicly available at \url{https://github.com/DanielTrosten/mvc}.

    \subsection{Details of pre-trained models}
      
For RGB-D, we use the following pre-trained models to extract features for the respective views:
\begin{itemize}
  \item View 1: ResNet-50 pre-trained on the ImageNet dataset. We use the version available in PyTorch\footnote{Documentation for the model can be found at \url{https://pytorch.org/docs/stable/torchvision/models.html}}, and remove the last (classification) layer.
  \item View 2: Doc2Vec pre-trained on the Wikipedia dataset. We use the pre-trained model available at \url{https://github.com/jhlau/doc2vec}.
\end{itemize}
Note that the same types of architectures are used in \cite{zhouEndtoEndAdversarialAttentionNetwork2020} to extract features for RGB-D. However, the authors do not supply the model and training details required to exactly reproduce their features.

    \subsection{Model architectures and hyperparameters}
      
\begin{table}
  \centering
  \begin{subfigure}{\columnwidth}
    \centering
    \begin{tabular}{cccc}\toprule
        Layer type & Neurons & Activation  & Batch-norm \\ \midrule
        \texttt{FC}         & 512     & \( \relu \) & \FALSE     \\
        \texttt{FC}         & 512     & \( \relu \) & \FALSE     \\
        \texttt{FC}         & 256     & \( \relu \) & \FALSE     \\
        \bottomrule
    \end{tabular}
    \caption{Fully connected encoder. \texttt{FC}: fully connected layer.}
    \label{tab:arch_mlp}
  \end{subfigure}
  \vspace{0.5cm}

  \begin{subfigure}{\columnwidth}
    \centering
    \bgroup
    \setlength{\tabcolsep}{4pt}
    \begin{tabular}{ccccc}\toprule
        \smalltable{Layer\\type} & \smalltable{Filter\\size}      & Filters & Activation  & \smalltable{Batch-\\norm} \\ \midrule
        \texttt{Conv}       & \( 5 \times 5 \) & 32         & \( \relu \) & \FALSE     \\
        \texttt{Conv}       & \( 5 \times 5 \) & 32         & \( \relu \) & \TRUE      \\
        \texttt{MaxPool}    & \( 2 \times 2 \) & --         & --          & \FALSE     \\
        \texttt{Conv}       & \( 3 \times 3 \) & 32         & \( \relu \) & \FALSE     \\
        \texttt{Conv}       & \( 3 \times 3 \) & 32         & \( \relu \) & \TRUE      \\
        \texttt{MaxPool}    & \( 2 \times 2 \) & --         & --          & \FALSE     \\
        \bottomrule
    \end{tabular}
    \egroup
    \caption{Convolutional neural network encoder. \texttt{Conv}: convolutional layer. \texttt{MaxPool}: max-pooling layer. Note that Batch normalization is applied before the activation function.}
    \label{tab:arch_cnn}
  \end{subfigure}
  \vspace{0.5cm}

  \begin{subfigure}{\columnwidth}
    \centering
    \begin{tabular}{cccc}\toprule
        Layer type & Neurons & Activation  & Batch-norm \\ \midrule
        FC         & 100     & \( \relu \) & \TRUE      \\
        FC         & \( k \) & softmax     & \FALSE     \\
        \bottomrule
    \end{tabular}
    \caption{Clustering module. \( k \) denotes the number of clusters.}
    \label{tab:arch_ddc}
  \end{subfigure}
  \caption{Network architectures.}
  \label{tab:arch}
\end{table}

\baseModel and \contrastiveModel trained on VOC, CCV, and RGB-D use fully connected encoders (Table \ref{tab:arch_mlp}) for all views. On E-MNIST, E-FMNIST and COIL our models use convolutional neural network encoders for all views (Table \ref{tab:arch_cnn}). The clustering module (Table \ref{tab:arch_ddc}) is the same for all experiments with \baseModel and \contrastiveModel.

Table \ref{tab:hyperparameters} lists the other hyperparameters that are not part of the model architectures. We use gradient clipping, and clip gradients with norms greater than "Max gradient norm". For some datasets, we found that decaying the learning rate helped the models converge. On these datasets, we reduce the learning rate once, at epoch "Decay step", with a factor of "Decay factor".

\begin{table*}
  \centering
  \bgroup
\setlength{\tabcolsep}{4.5pt}
\def\datasetsep{\midrule}
\begin{tabular}{lcccccccccc}\toprule
  Dataset                   & Model             & \smalltable{Batch\\size} & Epochs & \( \tau \) & \( \delta \) & \smalltable{Negative\\samples} & \smalltable{Max\\gradient \\norm} & \smalltable{Initial\\learning\\rate} & \smalltable{Decay\\step} & \smalltable{Decay\\factor} \\ \midrule
  \multirow{2}{*}{VOC}      & \baseModel        & \( 100 \)                & \( 100 \) & \( 0.1\)   & \( 0.1 \)    & \MIS                              & \( 5 \)                           & \( 0.001 \)                          & \( 50 \)                 & \( 0.1 \) \\
                            & \contrastiveModel & \( 100 \)                & \( 100 \) & \( 0.1\)   & \( 0.1 \)    & 25                                & \( 5 \)                           & \( 0.001 \)                          & \MIS                     & \MIS \\ \datasetsep
  \multirow{2}{*}{CCV}      & \baseModel        & \( 100 \)                & \( 100 \) & \( 0.1\)   & \( 20 \)     & \MIS                              & \( 5 \)                           & \( 0.001 \)                          & \MIS                     & \MIS \\
                            & \contrastiveModel & \( 100 \)                & \( 100 \) & \( 0.1\)   & \( 20 \)     & 25                                & \( 5 \)                           & \( 0.001 \)                          & \( 50 \)                 & \( 0.1 \) \\ \datasetsep
  \multirow{2}{*}{E-MNIST}  & \baseModel        & \( 100 \)                & \( 100 \) & \( 0.1\)   & \( 0.1 \)    & \MIS                              & \( 5 \)                           & \( 0.001 \)                          & \MIS                     & \MIS \\
                            & \contrastiveModel & \( 100 \)                & \( 100 \) & \( 0.1\)   & \( 0.1 \)    & 25                                & \( 5 \)                           & \( 0.001 \)                          & \MIS                     & \MIS \\ \datasetsep
  \multirow{2}{*}{E-FMNIST} & \baseModel        & \( 100 \)                & \( 100 \) & \( 0.1\)   & \( 0.1 \)    & \MIS                              & \( 5 \)                           & \( 0.001 \)                          & \MIS                     & \MIS \\
                            & \contrastiveModel & \( 100 \)                & \( 100 \) & \( 0.1\)   & \( 0.1 \)    & 25                                & \( 5 \)                           & \( 0.001 \)                          & \MIS                     & \MIS \\ \datasetsep
  \multirow{2}{*}{COIL-20}  & \baseModel        & \( 100 \)                & \( 100 \) & \( 0.1\)   & \( 20 \)     & \MIS                              & \( 5 \)                           & \( 0.001 \)                          & \MIS                     & \MIS \\
                            & \contrastiveModel & \( 100 \)                & \( 100 \) & \( 0.1\)   & \( 20 \)     & 25                                & \( 5 \)                           & \( 0.001 \)                          & \MIS                     & \MIS \\ \datasetsep
  \multirow{2}{*}{RGB-D}    & \baseModel        & \( 100 \)                & \( 100 \) & \( 0.1\)   & \( 0.1 \)    & \MIS                              & \( 5 \)                           & \( 0.001 \)                          & \MIS                     & \MIS \\
                            & \contrastiveModel & \( 100 \)                & \( 100 \) & \( 0.1\)   & \( 0.1 \)    & 25                                & \( 5 \)                           & \( 0.001 \)                          & \( 50 \)                 & \( 0.5 \) \\ \datasetsep
\end{tabular}
\egroup

  \caption{Hyperparameters used to train \baseModel and \contrastiveModel.}
  \label{tab:hyperparameters}
\end{table*}

    \subsection{Evaluation protocol}
      
For VOC, CCV, and E-MNIST, we use the baseline results obtained by \cite{zhouEndtoEndAdversarialAttentionNetwork2020}. For all baseline models, except EAMC, they run the model \( 10 \) times and report the average ACC and NMI. For EAMC, they train the model \( 20 \) times, and report the results from the run which resulted in the lowest value of the loss function. We follow the same evaluation procedure for EAMC, when we evaluate it on E-FMNIST, COIL-20, and RGB-D.

    \subsection{Experiments on NUS-WIDE-Animal}
      
\begin{table}
  \centering
  \setlength{\tabcolsep}{3pt}
  \begin{tabular}{ccccc} \toprule
    & SwMPC \cite{wangParameterFreeWeightedMultiView2020} & RSwMPC \cite{wangRobustSelfWeightedMultiView2020} & SiMVC & CoMVC \\ \midrule
    ACC & 0.1679 & 0.2778 & 0.2717 & \textbf{0.2892} \\
    NMI & 0.0899 & 0.1810 & 0.1767 & \textbf{0.2052} \\ \bottomrule
  \end{tabular}
  \vspace{0.3em}
  \caption{Comparison with \cite{wangParameterFreeWeightedMultiView2020,wangRobustSelfWeightedMultiView2020} on NUS-WIDE-Animal \cite{wangRobustSelfWeightedMultiView2020}.}
  \label{tab:results}
\end{table}
Table \ref{tab:results} shows the performance of our models on NUS-WIDE-Animal \cite{wangRobustSelfWeightedMultiView2020} (a subset of NUS-WIDE containing the animal classes only) compared to two additional models \cite{wangParameterFreeWeightedMultiView2020, wangRobustSelfWeightedMultiView2020} (as reported in \cite{wangRobustSelfWeightedMultiView2020}). SiMVC performs comparable, while CoMVC outperforms the competitors.

    \subsection{Training times}
      
\begin{table}
  \centering
  \begin{tabular}{cccc} \toprule
    & EAMC & SiMVC & CoMVC \\ \midrule
    sec/epoch & 33.48 & \textbf{12.18} & 14.28 \\
    \toprule
  \end{tabular}
  \caption{Time spent per training epoch for EAMC, SiMVC and CoMVC on E-FMNIST.}
  \label{tab:trainingTimes}
\end{table}

Table \ref{tab:trainingTimes} shows the average time spent per training epoch for EAMC, SiMVC, and CoMVC. Both SiMVC and CoMVC are more than twice as fast to train per epoch, when compared to EAMC. We believe that this is due to the extra components (attention network, discriminator) included in EAMC. SiMVC is a bit faster to train than CoMVC, due to the extra computations introduced by the contrastive loss.

\end{document}